\documentclass[letterpaper]{article} 
\usepackage{aaai2026}  
\usepackage{times}  
\usepackage{helvet}  
\usepackage{courier}  
\usepackage[hyphens]{url}  
\usepackage{graphicx} 
\usepackage{natbib}  
\usepackage{amsfonts}
\usepackage{amsmath}
\usepackage{mathrsfs}
\usepackage{caption} 
\usepackage{algorithm}
\usepackage{algorithmic}
\usepackage{float}
\usepackage{bm}
\usepackage{multirow}
\usepackage{graphicx}
\usepackage[table,xcdraw]{xcolor}
\usepackage{booktabs}
\usepackage{tabularx}
\usepackage{array}         
\usepackage{graphicx}      
\usepackage[table]{xcolor} 
\usepackage{subfig}
\usepackage{xcolor}
\usepackage{newfloat}
\usepackage{listings}
\DeclareCaptionStyle{ruled}{labelfont=normalfont,labelsep=colon,strut=off} 
\floatstyle{ruled}
\newfloat{listing}{tb}{lst}{}
\floatname{listing}{Listing}
\title{TimeMKG: Knowledge-Infused Causal Reasoning for \\ Multivariate Time Series Modeling}
\author{
    Yifei Sun\textsuperscript{\rm 1,2},
    Junming Liu\textsuperscript{\rm 2},
    Yirong Chen\textsuperscript{\rm 2},
    Xuefeng Yan\textsuperscript{\rm 1}\thanks{These authors contributed equally to this work as co-corresponding authors.},
    Ding Wang\textsuperscript{\rm 2}\footnotemark[1]
}
\affiliations {
    \textsuperscript{\rm 1}Key Laboratory of Smart Manufacturing in Energy Chemical Process, Ministry of Education, \\ East China University of Science and Technology, Shanghai, 200237, P. R. China\\
    \textsuperscript{\rm 2}Shanghai Artificial Intelligence Laboratory\\
    yifeisun@mail.ecust.edu.cn, xfyan@ecust.edu.cn, wangding@pjlab.org.cn
}

\usepackage{bibentry}

\begin{document}

\maketitle

\begin{abstract}
Multivariate time series data typically comprises two distinct modalities: variable semantics and sampled numerical observations. Traditional time series models treat variables as anonymous statistical signals, overlooking the rich semantic information embedded in variable names and data descriptions. However, these textual descriptors often encode critical domain knowledge that is essential for robust and interpretable modeling. Here we present TimeMKG, a multimodal causal reasoning framework that elevates time series modeling from low-level signal processing to knowledge informed inference. TimeMKG employs large language models to interpret variable semantics and constructs structured Multivariate Knowledge Graphs that capture inter-variable relationships. A dual-modality encoder separately models the semantic prompts—generated from knowledge graph triplets—and the statistical patterns from historical time series. Cross-modality attention aligns and fuses these representations at the variable level, injecting causal priors into downstream tasks such as forecasting and classification—providing explicit and interpretable priors to guide model reasoning. The experiment in diverse datasets demonstrates that incorporating variable-level knowledge significantly improves both predictive performance and generalization.
\end{abstract}


\section{Introdction}
\textbf{Multivariate time series data (MTSD)} are critical in domains such as industrial automation~\cite{Tii}, finance~\cite{finance}, energy systems~\cite{informer}, and healthcare~\cite{seong2024self}. However, their complex, high-dimensional and often non-stationary nature poses persistent challenges to modeling, interpretation, and decision making~\cite{TimeDP, STEM-LTS}. Traditional methods - ranging from statistical methods to modern deep learning~\cite{FCSTGNN} - have mainly focused on numerical temporal patterns, often overlooking the rich semantic information encoded in the names and descriptions of variables.

\begin{figure}[t]
\centering
\includegraphics[width=\columnwidth]{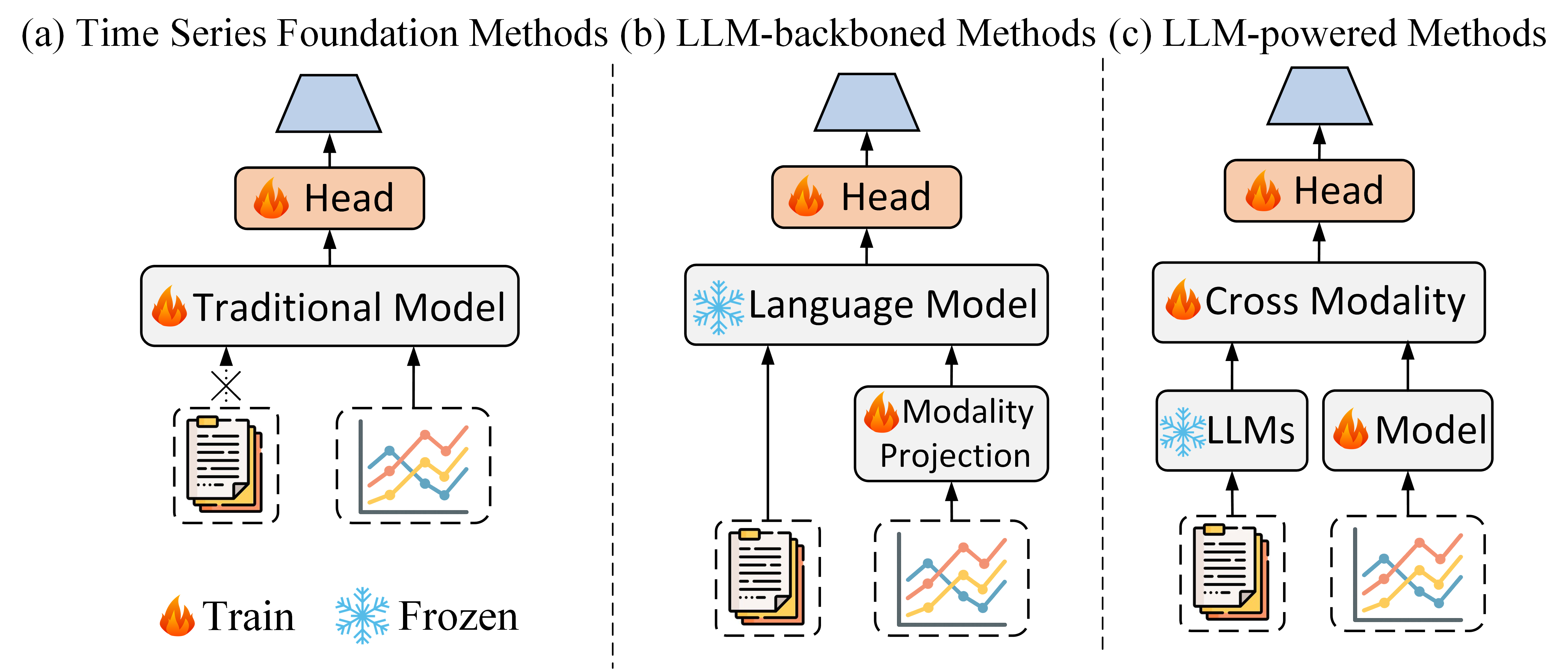}
\caption{(a) Time Series Foundation \& Traditional methods overlook header names and domain knowledge; (b) LLM-backboned methods tend to produce ambiguous semantic representations, as pretrained language models lack inherent temporal modeling capabilities; (c) LLM-powered methods integrate the strong semantic understanding of LLMs with the temporal modeling strengths of foundational models to enable more effective forecasting.}
\label{fig:introduction}
\end{figure}

However, MTSDs are inherently multimodal: the \textit{semantic information} conveyed by header names (textual modality) and the temporal dynamics captured in \textit{varying observations} (numerical modality). This dual structure naturally characterizes MTSD as a multimodal task. Textual descriptors often encode critical domain knowledge, such as causal relationships, physical meaning, and system roles, that can guide model reasoning~\cite{ChatTime}. Crucially, identical numerical values can have vastly different meanings across variables. For instance, a reading of 100 could represent normal engine temperature, but a life-threatening fever in medical contexts. Ignoring this semantic modality leads to blind spots in both interpretability and robustness, especially in safety-critical or knowledge-intensive applications.


Traditional time series models face significant challenges in leveraging causal information~\cite{casualtime}. On the one hand, statistical frameworks lack flexible mechanisms to encode prior domain knowledge~\cite{liXGboost}. On the other hand, most data-driven approaches~\cite{GMAN} often rely on task-specific architectures or hard-coded inductive biases to incorporate prior causal relationships among variables, making them difficult to generalize across tasks. As a result, modeling true causal interactions remains challenging, and models often depend on spurious correlations rather than grounded reasoning.


Large language models (LLMs)~\cite{qwen3} offer a powerful means of uncovering causal relationships between variables. Pretrained on large-scale text corpora, LLMs encode extensive common-sense and domain-specific knowledge, enabling them to interpret variable semantics and infer meaningful dependencies~\cite{Jin2023}. These capabilities make LLMs valuable tools for identifying latent relationships in MTSD. However, the knowledge they acquire remains implicit—embedded in opaque neural weights—lacking the transparency and structure needed for interpretability and refinement. This motivates the need for explicit, structured representations of causal knowledge.

In contrast, knowledge graphs (KGs) explicitly encode variable relationships as triplets, offering a clear and interpretable structure. KGs organize information through a clear entity–relation format, making them an ideal structure for storing causal relationships and highly suitable for injecting semantic priors into models. Susanti~\cite{susanti2024knowledge} showed that incorporating KGs as prompts allows LLMs to outperform fine-tuned models in few-shot causal discovery. Furthermore, Kim et al.~\cite{kim2024causal} demonstrated that graph-based prompts enhance LLMs’ understanding of causal chains via \textit{random walks}, significantly improving reasoning performance. KG is also conducive to expansion and dynamic updates: by simply adding or removing entities or edges, relevant knowledge can be instantly revised upon receiving new contexts.

In this work, we introduce \textbf{TimeMKG}, an LLMs-powered framework that leveraging KG for multivariate time series modeling. TimeMKG views MTSD as a dual-modality task, aligning: (1) The textual modality, where LLMs interpret variable names, descriptions, and expert knowledge to construct a causal knowledge graph; (2) The numerical modality, where statistical dependencies are extracted from temporal observations.
By aligning these modalities through cross-attention, TimeMKG injects causal priors from the semantic space into temporal modeling, enabling more interpretable and robust reasoning. Our contributions are as follows:

\begin{itemize}
    \item \textbf{Variable-level semantic modeling}: We are the first to introduce variable names as an input modality and explicitly incorporate a textual branch for MTSD modeling. Using the semantic understanding capabilities of LLMs, we extend domain-level causal inference to the fine-grained variable level, enabling consistent interpretation across datasets.
    \item \textbf{Explicit causal KG construction}: We automatically extract causal triplets between variables using LLMs to construct a KG that is human-auditable and incrementally updatable. This explicit structure significantly improves the performance of causal reasoning and prompt generation tasks.
    \item \textbf{LLM-powered dual-modality framework}: TimeMKG adopts a dual-branch design, LLMs extract causal relationships from textual semantics, while time-series models capture statistical dependencies in numerical observations. This unified framework supports diverse time series tasks while explicitly capturing causal dependencies.
    \item \textbf{SOTA performance on multiple tasks}: TimeMKG achieves consistently superior performance across long-term and short-term forecasting and classification tasks in diverse domains, with notable gains in interpretability and accuracy.
\end{itemize}

\section{Related Work}

Deep learning models have been widely explored for MTSD modeling. \textbf{CNN-based} methods~\cite{TCNs,Timesnet} extract intra and inter-variable features by applying sliding window convolutions. \textbf{RNN-based} methods~\cite{DARNN,deepar,N-BEATS} accumulate historical information through recurrent hidden states and alleviate the vanishing gradient problem using gated mechanisms. \textbf{Linear-based} models~\cite{TiDE,Dlinear} improve predictive performance by separately modeling trend and seasonal components. Early \textbf{Transformer-based} methods~\cite{informer} capture nonlinear dependencies across time steps via self-attention mechanisms. However, multivariate inputs are typically projected into a unified dimension at the embedding stage, sacrificing the preservation of variable-specific statistical properties.


Due to the sequential nature of similarities between textual and time series modality, recent LLMs advances have been extended to time series analysis. As shown in Figure \ref{fig:introduction}, these methods can be broadly categorized into three types:

(1) \textbf{Time Series Foundation Methods} are trained on large-scale time series data using strategies inspired by LLMs, aiming to learn generalizable temporal representations across diverse domains. Time-MoE~\cite{Time-MoE} introduces a sparse mixture-of-experts mechanism to reduce computational and inference costs. Moirai~\cite{Moirai} addresses cross-domain dimensional inconsistency by unfolding multivariate time series into a unified format during training. Although these models aim to solve zero-shot forecasting in the time series domain, their performance on private, domain-specific datasets remains limited.

(2) \textbf{LLM-backboned Methods}: To enable LLMs to process temporal information, numerical sequences must be discretized into univariate tokens and mapped into pseudo-word representations. For example, TimeLLM~\cite{TimeLLM} concatenates prompts with numerical tokens and feeds them into a frozen LLM, using the final hidden states as model outputs. However, since LLMs are pre-trained on natural language corpora rather than structured temporal data, their modeling effectiveness in this context has been questioned~\cite{areuse}.

(3) \textbf{LLM-powered Methods}: These methods~\cite{TimeCMA} adopt a dual-modality paradigm, where LLMs focus on the textual modality—leveraging their strong language understanding and generation capabilities—while standard time-series models handle temporal dependencies in numerical data. This “division-of-responsibility” framework enables complementary strengths: LLMs contribute semantic and causal insights, while temporal models capture statistical dynamics. This facilitates an organic fusion of semantic knowledge and temporal modeling.

\begin{figure*}[t]
\centering
\includegraphics[width=\textwidth]{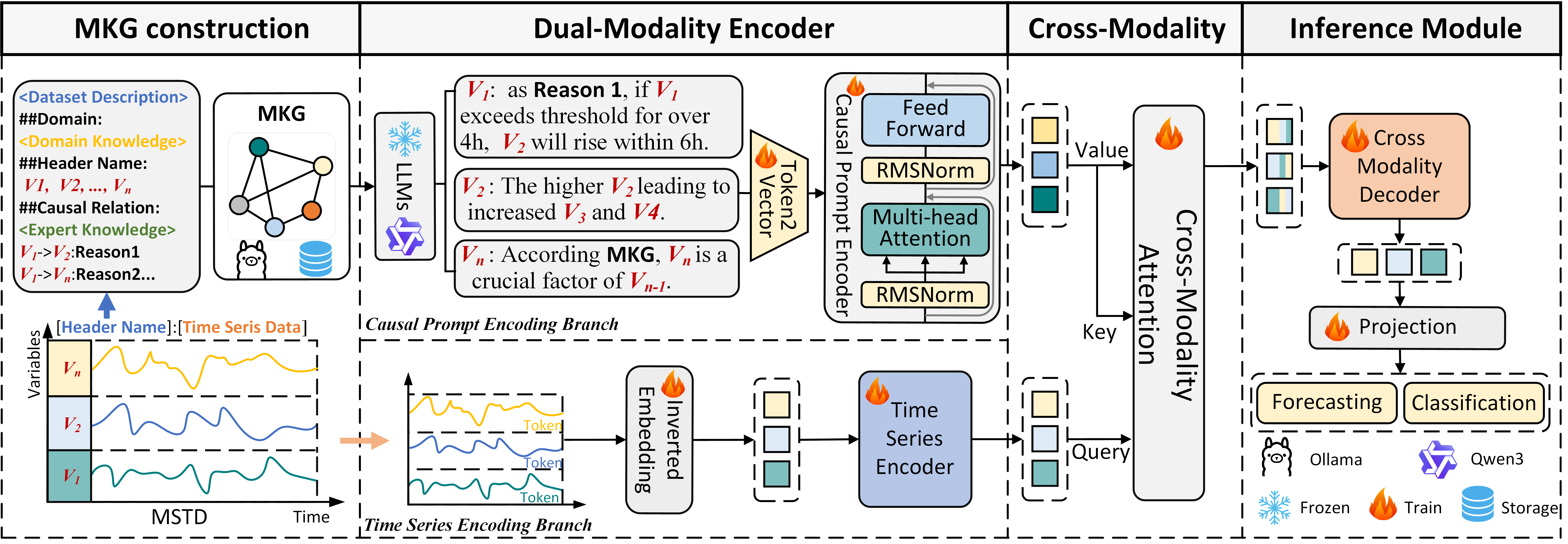}
\caption{Overall structure of TimeMKG. First, LLMs are used to extract causal knowledge and construct a knowledge graph. Then, a dual-modality encoder is used to model the prior causal relationship and statistical correlation between variables respectively. Finally, cross-modality attention is used to summarize the two types of representations and apply to the target tasks.}
\label{overall}
\end{figure*}

\section{Preliminaries}
\textbf{Multivariate Time Series Data}. Numerical modality is denoted as \({{\mathbf{X}} _{1:T}}=\left \{ {x}_{1},{x}_{2},\dots,{x}_{T} \right \}\in \mathbb{R}^{T\times N}\), with \(T\) time steps and $N$ variables. Textual modality \({\mathcal{V}}=\left \{  {v}_{1},{v}_{2},...,{v}_{N} \right \} \in \mathbb{R} ^{N} \) contains \(N\) header variable names.

\textbf{Knowledge Graph Definition}. Define a Graph \({\mathcal{G}}=\left ({\mathcal{V,E,R}}\right )\) to express semantic and causal relationships between variables. Where $\mathcal{V}$ denotes variable nodes, \({\mathcal{E}}\subseteq {\mathcal{V}}\times{\mathcal{R}}\times{\mathcal{V}}\) represents directed edges between them, \({\mathcal{R}}\) defines the relation types.

\textbf{Problem Definition}. Given a historical sequence \(\mathbf{X}_{1:T}\), \textbf{Forecasting} task is to predict the future $L$ time steps ${\mathbf{Y}_{T:T+L} }=\left \{ {x}_{T+1},...,{x}_{T+L} \right \} \in \mathbb{R}^{L\times N}$, ${f} _\mathrm{fore}: \mathbf{X}_{1:T}\mapsto\mathbf{Y}_{T:T+L}$.  \textbf{Classification} is to 
classify $\mathbf{X}_{1:T}$ into categories, ${f} _\mathrm{cls}: \mathbf{X}_{1:T}\mapsto\hat{c} \in \mathbb{R}^{C}$.

\section{Methodology}
\subsection{Overall Architecture}
As shown in Figure \ref{overall}, the overall structure of TimeMKG consists of four key modules: multivariate knowledge graph, dual-modality encoder, cross-modality attention, and inference module.

The multivariate knowledge graph is built by prompting a pretrained LLM with variable descriptions and causal knowledge to support domain reasoning. The Dual-Modality Encoder consists of causal prompt and time series branches, and has been demonstrated in TimeCMA~\cite{TimeCMA} to be effective for multimodal time series tasks. The causal branch uses a frozen LLM and a causal prompt encoder to refine inter-variable logic, while the time series branch employs inverted embeddings to capture inter-variable patterns from the numerical modality. Cross-Modality Attention aligns the two modalities into unified representations. The Inference Module decodes them via a cross-modality decoder and maps the result to task-specific outputs.

\subsection{Multivariate Knowledge Graph}
Recent advances~\cite{REBEL} in LLMs have shown remarkable capabilities in extracting entities, relationships, and structured patterns from unstructured data. Building on this strength, we propose the \textbf{M}ultivariate \textbf{K}nowledge \textbf{G}raph (\textbf{MKG}), a domain-specific graph representation tailored for MTSD. In MKG, nodes represent time-dependent variables, while directed edges capture causal or functional dependencies, serving as a prior graph to inform temporal modeling. 

To build the proposed MKG, we first refine the generated variable-level textual descriptions \(\hat{{\mathcal{S}}}\), optionally enriching them with external domain-specific knowledge \({\mathcal{T}}\), forming a comprehensive input for graph construction. The combined knowledge is fed into the LightRAG model~\cite{lightrag,aligning}, chosen for its efficiency in multi-hop reasoning and dynamic knowledge integration.

\begin{equation}
{{\mathcal{G}}_\mathcal{M}} = \text{LightRAG}\left( {\mathcal{I}} \right), \quad {\mathcal{I}} = \hat{\mathcal{S}} \oplus {\mathcal{T}},
\label{eq:mtkg_construction}
\end{equation}
Here, \(\oplus\) denotes optional concatenation with external textual knowledge \({\mathcal{T}}\). The combined input is then fed into LightRAG, which employs an LLM to infer graph structures that capture meaningful relationships among variables.

The obtained MKG, \({{\mathcal{G}}_\mathcal{M}} = ({\mathcal{V}}, {\mathcal{E}}, {\mathcal{R}})\), consists of a set of nodes \({\mathcal{V}}\) representing variables, a set of directed edges \({\mathcal{E}}\), and a corresponding set of relations \({\mathcal{R}}\). 

\begin{equation}
{{\mathcal{G}}_\mathcal{M}} = \left\{ ({v}_i,\ r,\ {v}_j)\ \middle|\ {v}_i, {v}_j \in {\mathcal{V}},\ r \in {\mathcal{R}} \right\},
\label{eq:MKG}
\end{equation}

Each edge \({e} \in {\mathcal{E}}\) is represented as a triplet \(({v}_i, {r}, {v}_j)\), where \({v}_i, {v}_j \in {\mathcal{V}}\) and \({r} \in {\mathcal{R}}\), indicating that a directed causal relation linking \({v}_i\) to \({v}_j\) via \({r}\). MKG construction leverages LLMs to automatically extract meaningful inter-variable relationships without manual annotation.

\subsection{Dual-Modality Encoder}
\subsubsection{Causal Prompt Encoding Branch.} Given an input header name \({v}_k\in {\mathcal{V}}\), we construct a variable-specific query \(\mathcal{Q} \left ( v_k \right ) \). Following LightRAG’s hybrid retrieval strategy, relevant triplets are extracted from \({{\mathcal{G}}_\mathcal{M}}\) by leveraging structural patterns during LLM inference.
\begin{equation}
{{\mathcal{G}}_\mathcal{M}}_{{v}_k} = \text{Retrieve}_{\text{global+local}}\left( \mathcal{Q}({v}_k),\ {{\mathcal{G}}_\mathcal{M}} \right),
\label{eq:hybrid_retrieval}
\end{equation}
where \(\text{Retrieve}_{\text{global}}\) captures broader semantic or functional groupings, while \(\text{Retrieve}_{\text{local}}\) targets direct relational dependencies of \(v_k\). The augmented prompt incorporates inter-variable causal evidence and domain knowledge of the dataset:
\begin{equation}
{p}_{{v}_k}=\mathcal{Q}({v}_k) \|(\bigcup_{{{\mathcal{G}}_\mathcal{M}}_{{v}_k}}[{v}_i] \rightarrow {r} \rightarrow[{v}_j]).
\label{eq:vk_prompt}
\end{equation}

Note that both the \({\mathcal{G}}_\mathcal{M}\) and the task-specific causal prompts \({\mathcal{P}}=\left \{  {p}_{1},{p}_{2},...,{p}_{N} \right \} \in \mathbb{R} ^{N\times l_{seq}} \) for the \(N\) variables are precomputed and stored in a database. This design significantly reduces redundant computation during training and inference. 

\subsubsection{Token2Vector.} To incorporate causal prompts into the TimeMKG, casual prompts \({\mathcal{P}}\) are first tokenized and truncated to a predefined maximum length \(l_{\text{max}}\), resulting in \({{\mathcal{P}}} \in \mathbb{R}^{N \times l_{\text{max}} \times D}\), where \(D\) is the embedding dimensions of the LLM. However, this high-dimensional representation is incompatible with the TimeMKG encoder, which requires a fixed-size vector per variable. To resolve this, the Token2Vector module maps each tokenized prompt into a compact vector via a two-layer perceptron.

\begin{equation}
\hat{{\mathcal{P}}} = \mathbf{W}_{\text{D2d}}\cdot \sigma(\mathbf{W}_{\text{pool}}{\mathcal{P}} + \mathbf{\beta}_{\text{pool}})+\mathbf{\beta}_{\text{D2d}}
\label{eq:demension}
\end{equation}
Here, \(\mathbf{W}_{\text{pool}} \in \mathbb{R}^{l_{\text{max}} \times 1}\) serves as a learnable pooling operator that aggregates token-level embeddings into a single vector of size \(D\), and \(\mathbf{W}_{\text{D2d}} \in \mathbb{R}^{D \times d}\) projects this vector into a lower-dimensional space \(\mathbb{R}^{d}\). The final output is \(\hat{\mathcal{P}} \in \mathbb{R}^{N \times d}\), where \(d\) is the target embedding dimension of TimeMKG. 

\subsubsection{Causal Prompt Encoder.} The processed prompts are then encoded by the causal prompt encoder \(\mathit{CPEncoder}(\cdot)\). We adopt the Pre-LN Transformer architecture to ensure training stability and efficient gradient flow. At the \(i_{th}\) layer, the causal embedding \(\hat{{\mathcal{P}}^i}\) is normalized via RMSNorm:
\begin{equation}
\tilde{{\mathcal{P}}^i}=\mathit{RN}(\hat{{\mathcal{P}}^i}) = \gamma \odot \frac{\hat{{\mathcal{P}}^i}}{\sqrt{\frac{1}{D} \sum_{k=1}^{D} (\hat{{\mathcal{P}}^i})_k^2}}
\label{eq:RN}
\end{equation}
where \(\tilde{{{\mathcal{P}}^i}}\) is the normalized embedding, \(\gamma\) is a learnable scaling parameter, and \(\odot\) denotes element-wise multiplication.

Subsequently, \(\tilde{\mathcal{P}^i}\) passes through the Multi-Head Self-Attention layer \(\mathit{MHSA}(\cdot)\) and is combined with \(\hat{\mathcal{P}^i}\) via a residual connection.
\begin{equation}
\bar {{\mathcal{P}}^i}=\mathit{MHSA}(\tilde{{\mathcal{P}}^i})+\hat{{\mathcal{P}}^i}
\label{eq:MHSA_RES}
\end{equation}
\begin{equation}
\mathit{MHSA}(\tilde{{\mathcal{P}}^i}) = \rho_o ( \mathit{Attention} ( \rho_q \tilde{{\mathcal{P}}^i}, \rho_k \tilde{{\mathcal{P}}^i}, \rho_v \tilde{{\mathcal{P}}^i}))
\label{eq:MHSA}
\end{equation}
where \(\bar{{\mathcal{P}}^i}\) represents the output after the residual connection, and \(\rho_o\), \(\rho_q\), \(\rho_k\), and \(\rho_v\) are the linear projection weight matrices. \(\mathit{MHSA}(\cdot)\) models the causal relationships among variables and enables effective information aggregation.

After normalization \(\vec{{\mathcal{P}}^i}=\mathit{RN}(\bar{{\mathcal{P}}^i})\), the attention embedding \(\vec{{\mathcal{P}}^i}\) is fed into the feed-forward network \(\mathit{FFN}(\cdot)\).
\begin{equation}
\dot{{\mathcal{P}}^i}=\mathit{FFN}(\vec{{\mathcal{P}}^i})+\bar{{\mathcal{P}}^i}
\label{eq:FFN_RES}
\end{equation}
\begin{equation}
\mathit{FFN}(\vec{\mathcal{P}}^i) = \rho_2\cdot \sigma( \rho_1\vec{\mathcal{P}}^i  + \beta_1  )+\beta_2
\label{eq:FFN}
\end{equation}
where \(\dot{{\mathcal{P}}^i}\in \mathbb{R}^{N \times d}\), \(\sigma\) denotes the activation function, \(\rho_1\) and \(\rho_2\) serve as the learnable weight matrices, \(\beta_1\) and \(\beta_2\) act as the bias vectors. Finally, \(\dot{\mathcal{P}}\) denotes the output of the \(\mathit{CPEncoder}(\cdot)\) after multiple layers of feature extraction.

\begin{table*}[t]
\renewcommand{\arraystretch}{1.0}
\setlength{\tabcolsep}{1mm}
{\small
\begin{tabular}{cc|cc|cc|cc|cc|cc|cc|cc|cc|cc}
\toprule
\multicolumn{2}{c|}{Method}                          & \multicolumn{2}{c|}{\textbf{TimeMKG}}                                                  & \multicolumn{2}{c|}{TimeCMA}                                                  & \multicolumn{2}{c|}{TimeLLM}                                           & \multicolumn{2}{c|}{UniTime}                                              & \multicolumn{2}{c|}{Time-MoE}                                      & \multicolumn{2}{c|}{PatchTST}                                             & \multicolumn{2}{c|}{iTransformer}                                          & \multicolumn{2}{c|}{TimesNet}                                                & \multicolumn{2}{c}{DLinear}                                         \\ \midrule
\multicolumn{2}{c|}{Metric}                          & MSE                                   & MAE                                   & MSE                                   & MAE                                   & MSE                               & MAE                                & MSE                               & MAE                                   & MSE                               & MAE                            & MSE                               & MAE                                   & MSE                                & MAE                                   & MSE                                     & MAE                                & MSE                                & MAE                            \\ \midrule
\multicolumn{1}{c|}{}                          & 96  & {\color[HTML]{FE0000} \textbf{0.373}} & {\color[HTML]{FE0000} \textbf{0.387}} & {\color[HTML]{FE0000} \textbf{0.373}} & {\color[HTML]{0000EE} {\underline{0.391}}}    & 0.398                             & 0.410                              & 0.397                             & 0.418                                 & 0.453                             & 0.463                          & 0.389                             & 0.412                                 & {\color[HTML]{0000EE} {\underline{0.379}}} & 0.400                                 & {\color[HTML]{0000EE} {\underline{0.379}}}      & 0.400                              & {\color[HTML]{333333} 0.396}       & 0.411                          \\
\multicolumn{1}{c|}{}                          & 192 & {\color[HTML]{FE0000} \textbf{0.426}} & {\color[HTML]{0000EE} {\underline{0.425}}}    & {\color[HTML]{0000EE} {\underline{0.427}}}    & {\color[HTML]{FE0000} \textbf{0.421}} & 0.451                             & 0.440                              & 0.434                             & 0.439                                 & 0.505                             & 0.537                          & 0.429                             & 0.432                                 & 0.449                              & 0.441                                 & {\color[HTML]{0000EE} {\underline{0.427}}}      & 0.432                              & 0.445                              & 0.440                          \\
\multicolumn{1}{c|}{}                          & 336 & {\color[HTML]{FE0000} \textbf{0.457}} & {\color[HTML]{FE0000} \textbf{0.441}} & {\color[HTML]{0000EE} {\underline{0.458}}}    & {\color[HTML]{0000EE} {\underline{0.448}}}    & 0.473                             & 0.451                              & 0.468                             & 0.457                                 & 0.543                             & 0.591                          & 0.478                             & 0.464                                 & 0.492                              & 0.465                                 & 0.478                                   & 0.464                              & 0.487                              & 0.465                          \\
\multicolumn{1}{c|}{}                          & 720 & {\color[HTML]{0000EE} {\underline{0.468}}}    & {\color[HTML]{FE0000} \textbf{0.454}} & {\color[HTML]{FE0000} \textbf{0.449}} & {\color[HTML]{0000EE} {\underline{0.460}}}    & 0.469                             & 0.470                              & 0.469                             & 0.477                                 & 0.621                             & 0.722                          & 0.522                             & 0.506                                 & 0.522                              & 0.504                                 & 0.522                                   & 0.506                              & 0.513                              & 0.510                          \\
\multicolumn{1}{c|}{\multirow{-5}{*}{\rotatebox[origin=c]{90}{ETTh1}}}   & avg & {\color[HTML]{0000EE} {\underline{0.431}}}    & {\color[HTML]{FE0000} \textbf{0.427}} & {\color[HTML]{FE0000} \textbf{0.427}} & {\color[HTML]{0000EE} {\underline{0.430}}}    & 0.448                             & 0.443                              & 0.442                             & 0.448                                 & 0.531                             & 0.578                          & 0.455                             & 0.454                                 & 0.460                              & 0.453                                 & 0.452                                   & 0.451                              & 0.460                              & 0.457                          \\ \midrule
\multicolumn{1}{c|}{}                          & 96  & {\color[HTML]{0000EE} {\underline{0.290}}}    & {\color[HTML]{FE0000} \textbf{0.336}} & {\color[HTML]{FE0000} \textbf{0.286}} & {\color[HTML]{FE0000} \textbf{0.336}} & 0.295                             & {\color[HTML]{0000EE} {\underline{0.345}}} & 0.296                             & {\color[HTML]{0000EE} {\underline{0.345}}}    & 0.508                             & 0.593                          & 0.321                             & 0.364                                 & 0.364                              & 0.395                                 & 0.329                                   & 0.370                              & 0.387                              & 0.434                          \\
\multicolumn{1}{c|}{}                          & 192 & {\color[HTML]{FE0000} \textbf{0.362}} & {\color[HTML]{FE0000} \textbf{0.386}} & {\color[HTML]{0000EE} {\underline{0.363}}}    & {\color[HTML]{0000EE} {\underline{0.387}}}    & 0.386                             & 0.399                              & 0.374                             & 0.394                                 & 0.685                             & 1.031                          & 0.406                             & 0.413                                 & 0.450                              & 0.442                                 & 0.393                                   & 0.409                              & 0.516                              & 0.504                          \\
\multicolumn{1}{c|}{}                          & 336 & {\color[HTML]{FE0000} \textbf{0.386}} & {\color[HTML]{FE0000} \textbf{0.417}} & {\color[HTML]{0000EE} {\underline{0.406}}}    & {\color[HTML]{0000EE} {\underline{0.421}}}    & 0.419                             & 0.429                              & 0.415                             & 0.427                                 & 0.882                             & 1.670                          & 0.444                             & 0.446                                 & 0.442                              & 0.445                                 & 0.471                                   & 0.468                              & 0.636                              & 0.568                          \\
\multicolumn{1}{c|}{}                          & 720 & {\color[HTML]{0000EE} {\underline{0.418}}}    & {\color[HTML]{FE0000} \textbf{0.438}} & {\color[HTML]{FE0000} \textbf{0.417}} & {\color[HTML]{FE0000} \textbf{0.438}} & 0.425                             & {\color[HTML]{0000EE} {\underline{0.442}}} & 0.425                             & 0.444                                 & 1.144                             & 2.673                          & 0.447                             & 0.457                                 & 0.445                              & 0.457                                 & 0.510                                   & 0.493                              & 0.885                              & 0.684                          \\
\multicolumn{1}{c|}{\multirow{-5}{*}{\rotatebox[origin=c]{90}{ETTh2}}}   & avg & {\color[HTML]{FE0000} \textbf{0.364}} & {\color[HTML]{FE0000} \textbf{0.394}} & {\color[HTML]{0000EE} {\underline{0.368}}}    & {\color[HTML]{0000EE} {\underline{0.396}}}    & 0.381                             & 0.404                              & 0.378                             & 0.403                                 & 0.805                             & 1.492                          & 0.404                             & 0.420                                 & 0.425                              & 0.435                                 & 0.426                                   & 0.435                              & 0.606                              & 0.548                          \\ \midrule
\multicolumn{1}{c|}{}                          & 96  & {\color[HTML]{FE0000} \textbf{0.303}} & {\color[HTML]{FE0000} \textbf{0.347}} & {\color[HTML]{0000EE} {\underline{0.312}}}    & {\color[HTML]{0000EE} {\underline{0.351}}}    & 0.359                             & 0.381                              & 0.322                             & 0.363                                 & 0.564                             & 0.730                          & 0.361                             & 0.377                                 & 0.347                              & 0.376                                 & 0.359                                   & 0.392                              & 0.364                              & 0.385                          \\
\multicolumn{1}{c|}{}                          & 192 & {\color[HTML]{0000EE} {\underline{0.362}}}    & {\color[HTML]{FE0000} \textbf{0.377}} & {\color[HTML]{FE0000} \textbf{0.361}} & {\color[HTML]{0000EE} {\underline{0.378}}}    & 0.383                             & 0.393                              & 0.366                             & 0.387                                 & 0.652                             & 0.863                          & 0.399                             & 0.395                                 & 0.382                              & 0.397                                 & 0.413                                   & 0.419                              & 0.396                              & 0.402                          \\
\multicolumn{1}{c|}{}                          & 336 & {\color[HTML]{FE0000} \textbf{0.391}} & {\color[HTML]{FE0000} \textbf{0.392}} & {\color[HTML]{0000EE} {\underline{0.392}}}    & {\color[HTML]{0000EE} {\underline{0.401}}}    & 0.416                             & 0.414                              & 0.398                             & 0.407                                 & 0.698                             & 0.929                          & 0.431                             & 0.416                                 & 0.442                              & 0.429                                 & 0.438                                   & 0.435                              & 0.426                              & 0.423                          \\
\multicolumn{1}{c|}{}                          & 720 & {\color[HTML]{0000EE} {\underline{0.462}}}    & {\color[HTML]{FE0000} \textbf{0.436}} & {\color[HTML]{FE0000} \textbf{0.453}} & {\color[HTML]{0000EE} {\underline{0.438}}}    & 0.483                             & 0.449                              & 0.454                             & 0.440                                 & 0.795                             & 1.095                          & 0.488                             & 0.448                                 & 0.486                              & 0.456                                 & 0.505                                   & 0.466                              & 0.483                              & 0.458                          \\
\multicolumn{1}{c|}{\multirow{-5}{*}{\rotatebox[origin=c]{90}{ETTm1}}}   & avg & {\color[HTML]{FE0000} \textbf{0.379}} & {\color[HTML]{FE0000} \textbf{0.388}} & {\color[HTML]{0000EE} {\underline{0.380}}}    & {\color[HTML]{0000EE} {\underline{0.392}}}    & 0.410                             & 0.409                              & 0.385                             & 0.399                                 & 0.677                             & 0.904                          & 0.420                             & 0.409                                 & 0.414                              & 0.415                                 & 0.429                                   & 0.428                              & 0.417                              & 0.417                          \\ \midrule
\multicolumn{1}{c|}{}                          & 96  & {\color[HTML]{0000EE} {\underline{0.175}}}    & {\color[HTML]{FE0000} \textbf{0.257}} & {\color[HTML]{FE0000} \textbf{0.173}} & {\color[HTML]{0000EE} {\underline{0.258}}}    & 0.193                             & 0.280                              & 0.183                             & 0.266                                 & 0.468                             & 0.480                          & 0.203                             & 0.286                                 & 0.203                              & 0.286                                 & 0.185                                   & 0.267                              & 0.208                              & 0.309                          \\
\multicolumn{1}{c|}{}                          & 192 & {\color[HTML]{FE0000} \textbf{0.235}} & {\color[HTML]{FE0000} \textbf{0.301}} & {\color[HTML]{0000EE} {\underline{0.238}}}    & {\color[HTML]{FE0000} \textbf{0.301}} & 0.257                             & 0.318                              & 0.251                             & 0.310                                 & 0.655                             & 0.913                          & 0.265                             & 0.323                                 & 0.265                              & 0.323                                 & 0.250                                   & {\color[HTML]{0000EE} {\underline{0.306}}} & 0.300                              & 0.377                          \\
\multicolumn{1}{c|}{}                          & 336 & {\color[HTML]{FE0000} \textbf{0.296}} & {\color[HTML]{FE0000} \textbf{0.336}} & {\color[HTML]{0000EE} {\underline{0.297}}}    & {\color[HTML]{0000EE} {\underline{0.338}}}    & 0.317                             & 0.353                              & 0.319                             & 0.351                                 & 0.886                             & 1.680                          & 0.312                             & 0.348                                 & 0.326                              & 0.360                                 & 0.318                                   & 0.348                              & 0.398                              & 0.440                          \\
\multicolumn{1}{c|}{}                          & 720 & {\color[HTML]{FE0000} \textbf{0.386}} & {\color[HTML]{FE0000} \textbf{0.392}} & {\color[HTML]{0000EE} {\underline{0.393}}}    & {\color[HTML]{0000EE} {\underline{0.394}}}    & 0.419                             & 0.411                              & 0.420                             & 0.410                                 & 1.194                             & 2.894                          & 0.411                             & 0.403                                 & 0.424                              & 0.412                                 & 0.425                                   & 0.409                              & 0.565                              & 0.531                          \\
\multicolumn{1}{c|}{\multirow{-5}{*}{\rotatebox[origin=c]{90}{ETTm2}}}   & avg & {\color[HTML]{FE0000} \textbf{0.273}} & {\color[HTML]{FE0000} \textbf{0.322}} & {\color[HTML]{0000EE} {\underline{0.275}}}    & {\color[HTML]{0000EE} {\underline{0.323}}}    & 0.297                             & 0.341                              & 0.293                             & 0.334                                 & 0.800                             & 1.492                          & 0.298                             & 0.340                                 & 0.304                              & 0.345                                 & 0.294                                   & 0.333                              & 0.368                              & 0.414                          \\ \midrule
\multicolumn{1}{c|}{}                          & 96  & {\color[HTML]{FE0000} \textbf{0.162}} & {\color[HTML]{FE0000} \textbf{0.208}} & {\color[HTML]{0000EE} {\underline{0.167}}}    & {\color[HTML]{0000EE} {\underline{0.211}}}    & 0.198                             & 0.235                              & 0.171                             & 0.214                                 & 0.281                             & 0.243                          & 0.200                             & 0.238                                 & 0.211                              & 0.252                                 & 0.195                                   & 0.235                              & 0.200                              & 0.263                          \\
\multicolumn{1}{c|}{}                          & 192 & {\color[HTML]{FE0000} \textbf{0.210}} & {\color[HTML]{FE0000} \textbf{0.244}} & {\color[HTML]{0000EE} {\underline{0.212}}}    & {\color[HTML]{0000EE} {\underline{0.253}}}    & 0.240                             & 0.269                              & 0.217                             & 0.254                                 & 0.361                             & 0.338                          & 0.245                             & 0.275                                 & 0.255                              & 0.285                                 & 0.241                                   & 0.272                              & 0.239                              & 0.300                          \\
\multicolumn{1}{c|}{}                          & 336 & {\color[HTML]{FE0000} \textbf{0.260}} & {\color[HTML]{FE0000} \textbf{0.284}} & {\color[HTML]{0000EE} {\underline{0.270}}}    & {\color[HTML]{0000EE} {\underline{0.292}}}    & 0.295                             & 0.308                              & 0.274                             & 0.293                                 & 0.436                             & 0.441                          & 0.296                             & 0.310                                 & 0.306                              & 0.320                                 & 0.293                                   & 0.309                              & 0.284                              & 0.335                          \\
\multicolumn{1}{c|}{}                          & 720 & {\color[HTML]{FE0000} \textbf{0.327}} & {\color[HTML]{0000EE} {\underline{0.346}}}    & 0.350                                 & 0.348                                 & 0.368                             & 0.353                              & 0.351                             & {\color[HTML]{FE0000} \textbf{0.343}} & 0.541                             & 0.610                          & 0.369                             & 0.355                                 & 0.377                              & 0.365                                 & 0.367                                   & 0.356                              & {\color[HTML]{0000EE} {\underline{0.347}}} & 0.384                          \\
\multicolumn{1}{c|}{\multirow{-5}{*}{\rotatebox[origin=c]{90}{Weather}}} & avg & {\color[HTML]{FE0000} \textbf{0.240}} & {\color[HTML]{FE0000} \textbf{0.271}} & {\color[HTML]{0000EE} {\underline{0.250}}}    & {\color[HTML]{0000EE} {\underline{0.276}}}    & 0.275                             & 0.291                              & 0.253                             & 0.276                                 & 0.405                             & 0.408                          & 0.277                             & 0.294                                 & 0.287                              & 0.306                                 & 0.274                                   & 0.293                              & 0.268                              & 0.321                          \\ \midrule
\multicolumn{1}{c|}{}                          & 24  & {\color[HTML]{FE0000} \textbf{1.985}} & {\color[HTML]{0000EE} {\underline{0.965}}}    & {\color[HTML]{0000EE} {\underline{1.996}}}    & 0.998                                 & 2.383                             & 1.004                              & 2.460                             & 0.954                                 & 3.650                             & 1.245                          & 2.344                             & {\color[HTML]{FE0000} \textbf{0.926}} & 2.284                              & {\color[HTML]{FE0000} \textbf{0.926}} & 2.569                                   & 0.957                              & 4.541                              & 1.605                          \\
\multicolumn{1}{c|}{}                          & 36  & {\color[HTML]{FE0000} \textbf{1.902}} & {\color[HTML]{FE0000} \textbf{0.898}} & {\color[HTML]{0000EE} {\underline{1.906}}}    & 0.915                                 & 2.390                             & 0.993                              & 1.998                             & 0.912                                 & 4.320                             & 1.369                          & 2.410                             & 0.954                                 & 2.062                              & 0.932                                 & 2.046                                   & 0.899                              & 4.221                              & 1.508                          \\
\multicolumn{1}{c|}{}                          & 48  & {\color[HTML]{FE0000} \textbf{1.860}} & {\color[HTML]{FE0000} \textbf{0.859}} & {\color[HTML]{0000EE} {\underline{1.867}}}    & {\color[HTML]{0000EE} {\underline{0.868}}}    & 2.394                             & 1.003                              & 1.979                             & 0.912                                 & 4.737                             & 1.461                          & 2.250                             & 0.928                                 & 2.229                              & 0.965                                 & 2.248                                   & 0.921                              & 4.028                              & 1.453                          \\
\multicolumn{1}{c|}{}                          & 60  & {\color[HTML]{FE0000} \textbf{1.916}} & {\color[HTML]{0000EE} {\underline{0.910}}}    & {\color[HTML]{0000EE} {\underline{1.920}}}    & {\color[HTML]{FE0000} \textbf{0.904}} & 2.562                             & 1.049                              & 2.109                             & 0.938                                 & 5.197                             & 1.547                          & 2.175                             & 0.940                                 & 2.149                              & 0.952                                 & 2.124                                   & 0.919                              & 4.169                              & 1.460                          \\
\multicolumn{1}{c|}{\multirow{-5}{*}{\rotatebox[origin=c]{90}{ILI}}}     & avg & {\color[HTML]{FE0000} \textbf{1.916}} & {\color[HTML]{FE0000} \textbf{0.908}} & {\color[HTML]{0000EE} {\underline{1.922}}}    & {\color[HTML]{0000EE} {\underline{0.921}}}    & 2.432                             & 1.012                              & 2.137                             & 0.929                                 & 4.476                             & 1.405                          & 2.295                             & 0.937                                 & 2.181                              & 0.944                                 & 2.247                                   & 0.924                              & 4.240                              & 1.506                          \\ \midrule
\multicolumn{2}{c|}{1st/2nd}                         & {\color[HTML]{FE0000} \textbf{38}}    & {\color[HTML]{0000EE} {\underline{10}}}       & {\color[HTML]{FE0000} \textbf{13}}    & {\color[HTML]{0000EE} {\underline{32}}}       & {\color[HTML]{FE0000} \textbf{0}} & {\color[HTML]{0000EE} {\underline{2}}}     & {\color[HTML]{FE0000} \textbf{1}} & {\color[HTML]{0000EE} {\underline{3}}}        & {\color[HTML]{FE0000} \textbf{0}} & {\color[HTML]{0000EE} {\underline{0}}} & {\color[HTML]{FE0000} \textbf{1}} & {\color[HTML]{0000EE} {\underline{0}}}        & {\color[HTML]{FE0000} \textbf{1}}  & {\color[HTML]{0000EE} {\underline{1}}}        & {\color[HTML]{FE0000} { \textbf{0}}} & {\color[HTML]{0000EE} {\underline{3}}}     & {\color[HTML]{FE0000} \textbf{0}}  & {\color[HTML]{0000EE} {\underline{1}}} \\ \bottomrule
\end{tabular}
}
\caption{Long-term forecasting task. The historical sequence length \(T\) is set as 36 for ILI and 96 for the others. The prediction lengths \(L \in \left \{  24,36,48,60\right \}\) for ILI and \(L \in \left \{  96,192,336,720\right \}\) for the others. The best results highlighted in {\color[HTML]{FE0000} \textbf{bold}} and the second-best {\color[HTML]{0000EE} {\underline{underlined}}}. }
\label{table1:long term forecasting}
\end{table*}

\subsubsection{Time Series Encoding Branch.} In MTSD, the numerical modality input \({{\mathbf{X}}_{1:T}} \in \mathbb{R}^{T \times N}\) spans both temporal (\(T\)) and variable (\(N\)) dimensions. Traditional methods focus on \(T\) in \(\mathit{MHSA}(\cdot)\), generating a self-attention score matrix \(\mathbf{S} \in \mathbb{R}^{T \times T}\), while projecting the variable dimension \(N\) into a shared latent space. This, however, emphasizes temporal patterns but overlooks inter-variable dependencies. To address this limitation, we adopt an inverted embedding strategy~\cite{iTransformer} that treats variables as tokens to explicitly captures their statistical correlations.

\subsubsection{Inverted Embedding.} 
To treat variables as tokens, we first transpose the input to obtain a variable-oriented representation: 
\(\hat{\mathbf{X}} = \mathbf{X}_{1:T}^\top \in \mathbb{R}^{N \times T}\). 
Each variable \(i\) corresponds to a temporal sequence of length \(T\), denoted as \(\hat{\mathbf{X}}_i \in \mathbb{R}^T\). 
We embed each temporal sequence into a unified latent space using the following transformation:
\begin{equation}
\tilde{\mathbf{X}}_i = \mathbf{W}_{\text{emb}} \hat{\mathbf{X}}_i + \boldsymbol{\beta}_{\text{emb}}
\label{eq:embed}
\end{equation}
where \(\mathbf{W}_{\text{emb}}\) and \(\boldsymbol{\beta}_{\text{emb}}\) are learnable projection weights and bias terms, respectively. 
\(\tilde{\mathbf{X}} \in \mathbb{R}^{N \times d}\) denotes the embedded representation of all variables in the shared latent space.

\subsubsection{Time Series Encoder.}
Similar to the \(\mathit{CPEncoder}(\cdot)\), the Time Series Encoder \(\mathit{TSEncoder}(\cdot)\) adopts a Pre-LN Transformer to capture pairwise similarities among variables.
\begin{equation}
\bar {{\mathbf{X}}^i}=\mathit{MHSA}(\mathit{RN}(\tilde{{\mathbf{X}}^i}))+\tilde{{\mathbf{X}}^i}
\label{eq:mhsa_tse}
\end{equation}
\begin{equation}
\dot {{\mathbf{X}}^i}=\mathit{FFN}(\mathit{RN}(\bar{{\mathbf{X}}^i}))+\bar{{\mathbf{X}}^i}
\label{eq:ffn_ts}
\end{equation}
where the attention weight matrix \(\mathbf{S} \in \mathbb{R}^{N \times N}\) captures the relative importance among variables. The output \(\dot{\mathbf{X}} \in \mathbb{R}^{N \times d}\), obtained after \(\mathit{MHSA}(\cdot)\) and \(\mathit{FFN}(\cdot)\) with residual connections, effectively encodes the statistical correlations within the numerical modality.

\subsection{Cross-Modality Attention}
To integrate the numerical modality of MTSD with the textual modality embedded from the MKG, we employ cross-modality attention \(\mathit{CMA}(\cdot)\) to align modalities at the variable level. In TimeMKG, the numerical embedding \(\dot{\mathbf{X}}\in \mathbb{R}^{N \times d}\) is used as the \textit{Query}, while the textual embedding \(\dot{\mathcal{P}} \in \mathbb{R}^{N \times d}\) serves as the \textit{Key} and \textit{Value}.

\begin{equation}
\mathbf{S}_{N}=\mathit{softmax}\left(\frac{\mathbf{W}_q\dot{\mathcal{P}} \otimes  \mathbf{W}_k\dot{\mathbf{X}}^\top}{\sqrt{d}}\right)
\label{eq:cross_attention}
\end{equation}
Linear layers \(\mathbf{W}_q\), \(\mathbf{W}_k\), and \(\mathbf{W}_v\) are applied along the variable dimension, followed by dot-product similarity to compute the cross-modality score \(\mathbf{S}_{N} \in \mathbb{R}^{N \times N}\), which is scaled by \(\frac{1}{\sqrt{d}}\) to stabilize the values.

\begin{equation}
\mathbf{H}_{N} = \mathbf{W}_o(\mathbf{S}_{N} \otimes (\mathbf{W}_v {\dot{\mathcal{P}}}))+\dot{\mathbf{X}}
\label{eq:cross_fusion}
\end{equation}
\(\mathbf{S}_{N}\) is then used to aggregate the textual modality \(\dot{\mathcal{P}}\), allowing each numerical variable to attend to its semantically relevant counterparts in the MKG embeddings. Finally, a residual connection is applied to obtain the enhanced numerical representation \(\mathbf{H}_{N}\in \mathbb{R}^{N \times d}\).
\subsection{Inference Module}
We adopt a Transformer-based Cross Modality Decoder \(\mathit{CMD}(\cdot)\) to decode the fused cross-modality causal embeddings learned by the dual-modality encoder. This decoder captures both inter-variable and intra-variable dependencies, as well as modality interactions, crucial for downstream inference. A task-specific projection head maps the decoder output to the target space.

\section{Experiments}

\subsubsection{Datasets.}
To verify the effectiveness of TimeMKG in various tasks, we conduct extensive experiments on three mainstream tasks, including long-term and short-term forecasting, and classification. We evaluate TimeMKG across 16 diverse MTSD datasets: \textit{ETTh1}, \textit{ETTh2}, \textit{ETTm1}, \textit{ETTm2}, \textit{Weather}, and \textit{ILI} are used for long-term forecasting; \textit{ICL}, \textit{IoTFlow}, \textit{Nasdq}, \textit{Internet} and \textit{Battery} for short-term forecasting; \textit{SCP1}, \textit{SCP2}, \textit{Ethanol}, \textit{Heart}, and \textit{PEMS-SF} for classification. Detailed dataset descriptions and statistics are provided in the Appendix A.

\subsubsection{Baselines.}
We select 8 SOTA models spanning 6 categories: (1) \textbf{LLM-powered:} TimeCMA~\cite{TimeCMA}; (2) \textbf{LLM-backboned:} Time-LLM~\cite{TimeLLM}, UniTime~\cite{UniTime}; (3) \textbf{Time Series Foundation:} Time-MoE~\cite{Time-MoE}; (4) \textbf{Transformer-based:} PatchTST~\cite{PatchTST}, iTransformer~\cite{iTransformer}; (5) \textbf{TCN-based:} TimesNet~\cite{Timesnet}; (6) \textbf{Linear-based:} DLinear~\cite{Dlinear}. We further compare SOTA models for specific tasks, such as TimeXer~\cite{TimeXer} and FEDformer~\cite{fedformer} for short-term forecasting, XGBoost~\cite{XGBoost} and LSTNet~\cite{LSTNet} for classification, and so on (listed in Figure \ref{Figuire:classification})~\cite{DTW,LSTM,lightTS,TCNs}. In total, we compare TimeMKG against more than 20 baseline models across different categories.

\begin{figure}[t]
\centering
\includegraphics[width=\linewidth]{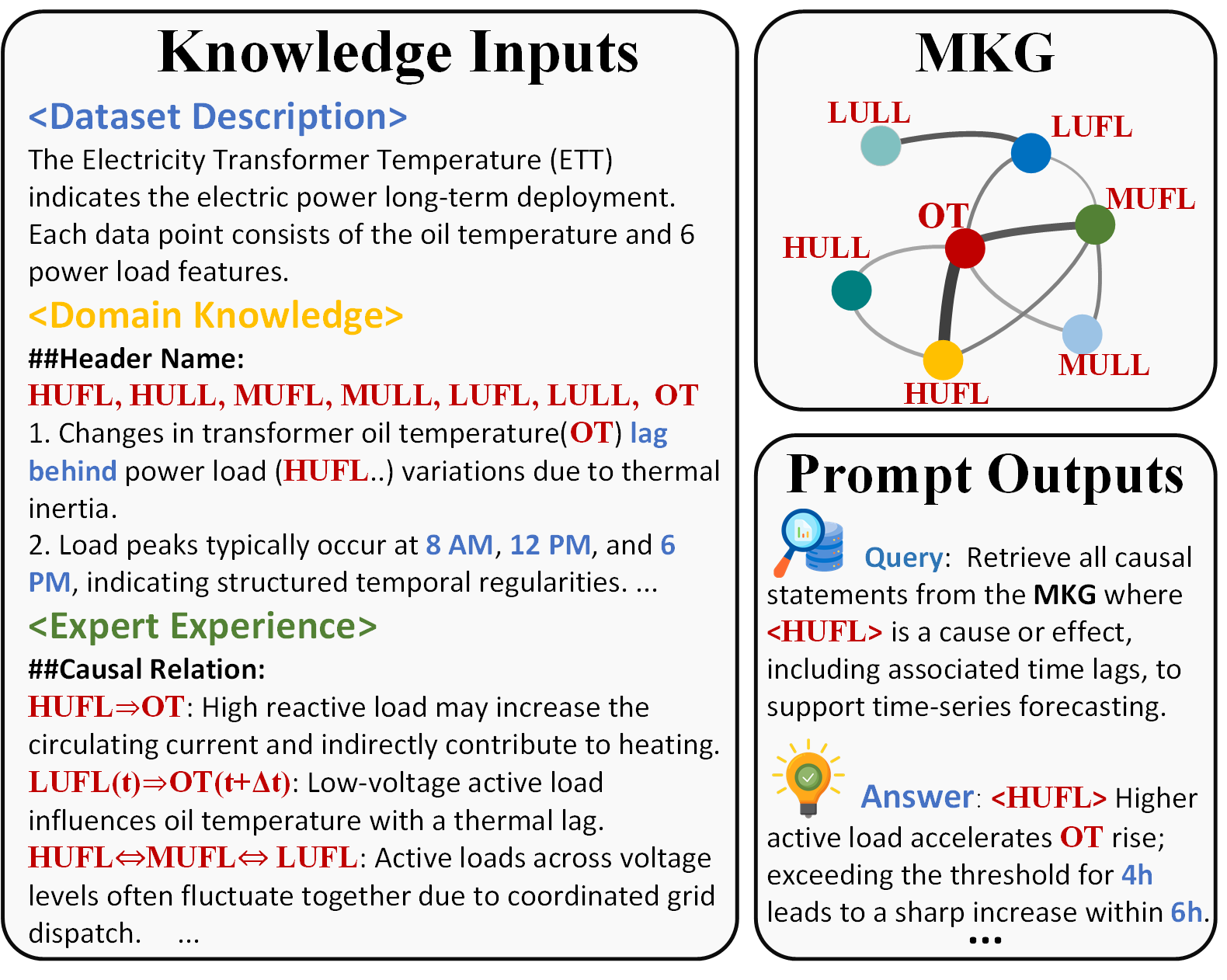}
\caption{An Example of MKG Construction for the ETT Datasets. \({\color{red} \mathbf{<>} } \) represents a variable-specific identifier, unique to each variable.}
\label{Figuire:Graphprompt}
\end{figure}

\subsubsection{Evaluation.}
We follow the evaluation protocol established in TimesNet~\cite{Timesnet}. Long-term forecasting: MSE and MAE; Short-term forecasting: SMAPE, MASE and OWA; Classification: Accuracy. All tested methods use the same test batch size to ensure fairness. Each experiment was repeated at least three times using different seeds on a single NVIDIA A100 GPU.

\subsubsection{MKG Example.} Recent studies suggest that textual LLMs are relatively insensitive to numerical patterns in time series, rendering sample-specific prompt designs (e.g., TimeLLM~\cite{TimeLLM}) less effective for modeling real-time dynamics. To overcome this limitation, TimeMKG focuses on inferring causal and semantic relationships among variables and stores knowledge in an explicit MKG. Figure~\ref{Figuire:Graphprompt} illustrates the construction of MKG and prompt design for the ETT dataset. Specifically, \textit{Dataset Description} provides contextual information about the application domain, \textit{Domain Knowledge} defines the graph nodes (i.e., variable headers) along with relevant public knowledge retrieved by LLMs, and \textit{Expert Experience}, optional and incorporated when available, encodes private domain-specific insights. Given a variable query, LLMs extract a subgraph from the MKG to generate prompt outputs. This structure enables the integration of both public and confidential domain knowledge, making TimeMKG well-suited for MTSD tasks with clearly defined header names.

\subsubsection{Long and Short-term Forecasting Tasks.} 
As shown in Table~\ref{table1:long term forecasting}, (1) TimeMKG consistently outperforms all baselines across datasets, achieving the best performance in 38/48 in Long-term sub-tasks and all short-term average tasks. (2) LLMs-based methods outperform conventional deep learning models, highlighting the benefits of semantic-aware multimodal modeling for MTSD tasks. (3) LLMs-powered methods (e.g., TimeMKG) surpass LLMs-backboned ones (e.g., TimeLLM and UniTime), with average improvements of 15\% and 7.3\% in MSE, respectively, indicating that the pretrained LLMs are better suited for textual modalities than serving as backbones. (4) By integrating MKG and causal knowledge, TimeMKG achieves superior performance on datasets with clearer variable semantics (e.g., Weather and ICL vs. ETT and IoTFlow), with an average improvement of 4\% and 8.9\% in SOTA metrics compared to the second-best methods, respectively.

\begin{table}[t]
\renewcommand{\arraystretch}{1.0}
\setlength{\tabcolsep}{0.65mm}
{\small
\begin{tabular}{cc|ccccccc}
\toprule
\multicolumn{2}{c|}{Method}                             & \small{\textbf{T}MKG} & \textbf{T}Xer & Patch* & iTrans* & FED*  & \textbf{T}sNet & DLinear \\ \midrule
\multicolumn{1}{c|}{}                           & \small{SMAPE} & {\color[HTML]{FE0000} \textbf{38.79}} & 43.77 & {\color[HTML]{0000EE} {\underline{42.59}}} & 44.04 & 71.26 & 47.81 & 72.24 \\
\multicolumn{1}{c|}{}                           & \small{MASE}  & {\color[HTML]{FE0000} \textbf{0.353}} & 0.422 & {\color[HTML]{0000EE} {\underline{0.391}}} & 0.411 & 0.725 & 0.461 & 0.790 \\
\multicolumn{1}{c|}{\multirow{-3}{*}{\rotatebox[origin=c]{90}{ICL}}}      & \small{OWA}   & {\color[HTML]{FE0000} \textbf{0.456}} & 0.527 & {\color[HTML]{0000EE} {\underline{0.502}}} & 0.505 & 0.880 & 0.577 & 0.921 \\ \midrule
\multicolumn{1}{c|}{}                           & \small{SMAPE} & {\color[HTML]{FE0000} \textbf{84.78}} & 85.15 & 84.91 & {\color[HTML]{0000EE} {\underline{84.82}}} & 90.92 & 84.85 & 88.22 \\
\multicolumn{1}{c|}{}                           & \small{MASE}  & {\color[HTML]{FE0000} \textbf{1.053}} & 1.056 & 1.056 & {\color[HTML]{0000EE} {\underline{1.055}}} & 1.181 & 1.058 & 1.316 \\
\multicolumn{1}{c|}{\multirow{-3}{*}{\rotatebox[origin=c]{90}{IoTFlow}}}  & \small{OWA}   & {\color[HTML]{FE0000} \textbf{1.542}} & 1.548 & 1.545 & {\color[HTML]{0000EE} {\underline{1.543}}} & 1.675 & 1.545 & 1.695 \\ \midrule
\multicolumn{1}{c|}{}                           & \small{SMAPE} & {\color[HTML]{FE0000} \textbf{22.51}} & 23.55 & 22.99 & {\color[HTML]{0000EE} {\underline{23.06}}} & 26.44 & 25.32 & 27.86 \\
\multicolumn{1}{c|}{}                           & \small{MASE}  & {\color[HTML]{FE0000} \textbf{3.136}} & 3.349 & {\color[HTML]{0000EE} {\underline{3.214}}} & 3.221 & 3.986 & 3.763 & 4.503 \\
\multicolumn{1}{c|}{\multirow{-3}{*}{\rotatebox[origin=c]{90}{Nasdq}}}    & \small{OWA}   & {\color[HTML]{FE0000} \textbf{1.116}} & 1.184 & {\color[HTML]{0000EE} {\underline{1.143}}} & {\color[HTML]{0000EE} 1.146} & 1.380 & 1.307 & 1.509 \\ \midrule
\multicolumn{1}{c|}{}                           & \small{SMAPE} & {\color[HTML]{FE0000} \textbf{62.24}} & 62.88 & 63.25 & {\color[HTML]{0000EE} {\underline{62.37}}} & 64.04 & 62.78 & 66.21 \\
\multicolumn{1}{c|}{}                           & \small{MASE}  & {\color[HTML]{FE0000} \textbf{0.838}} & {\color[HTML]{0000EE} {\underline{0.843}}} & 0.855 & 0.854 & 0.860 & 0.845 & 0.898 \\
\multicolumn{1}{c|}{\multirow{-3}{*}{\rotatebox[origin=c]{90}{Internet}}} & \small{OWA}   & {\color[HTML]{FE0000} \textbf{1.017}} & {\color[HTML]{0000EE} {\underline{1.027}}} & 1.036 & 1.028 & 1.046 & {\color[HTML]{0000EE} {\underline{1.027}}} & 1.086 \\ \midrule
\multicolumn{1}{c|}{}                           & \small{SMAPE} & {\color[HTML]{FE0000} \textbf{3.898}} & 4.712 & {\color[HTML]{0000EE} {\underline{3.907}}} & 4.611 & 12.15 & 4.156 & 19.44 \\
\multicolumn{1}{c|}{}                           & \small{MASE}  & {\color[HTML]{FE0000} \textbf{7.186}} & 8.922 & {\color[HTML]{0000EE} {\underline{7.232}}} & 8.739 & 22.14 & 7.752 & 32.77 \\
\multicolumn{1}{c|}{\multirow{-3}{*}{\rotatebox[origin=c]{90}{Battery}}}  & \small{OWA}   & {\color[HTML]{FE0000} \textbf{0.958}} & 1.197 & {\color[HTML]{0000EE} {\underline {0.959}}} & 1.167 & 3.167 & 1.039 & 5.075 \\ \bottomrule
\end{tabular}}
\caption{Short-term forecasting task. The historical sequence length \(T\) is set as {24}. All the results are averaged from 4 different prediction lengths \(L \in \left \{  12,18,24,36\right \}\). Lower metres values indicate more accurate predict. “\textbf{T}” indicates the name of Time. “*” in Transformers indicates the name of former.}
\label{short term forecasting}
\end{table}

\subsubsection{Classification Tasks.} 
\begin{figure}[ht]
\centering
\includegraphics[width=\linewidth]{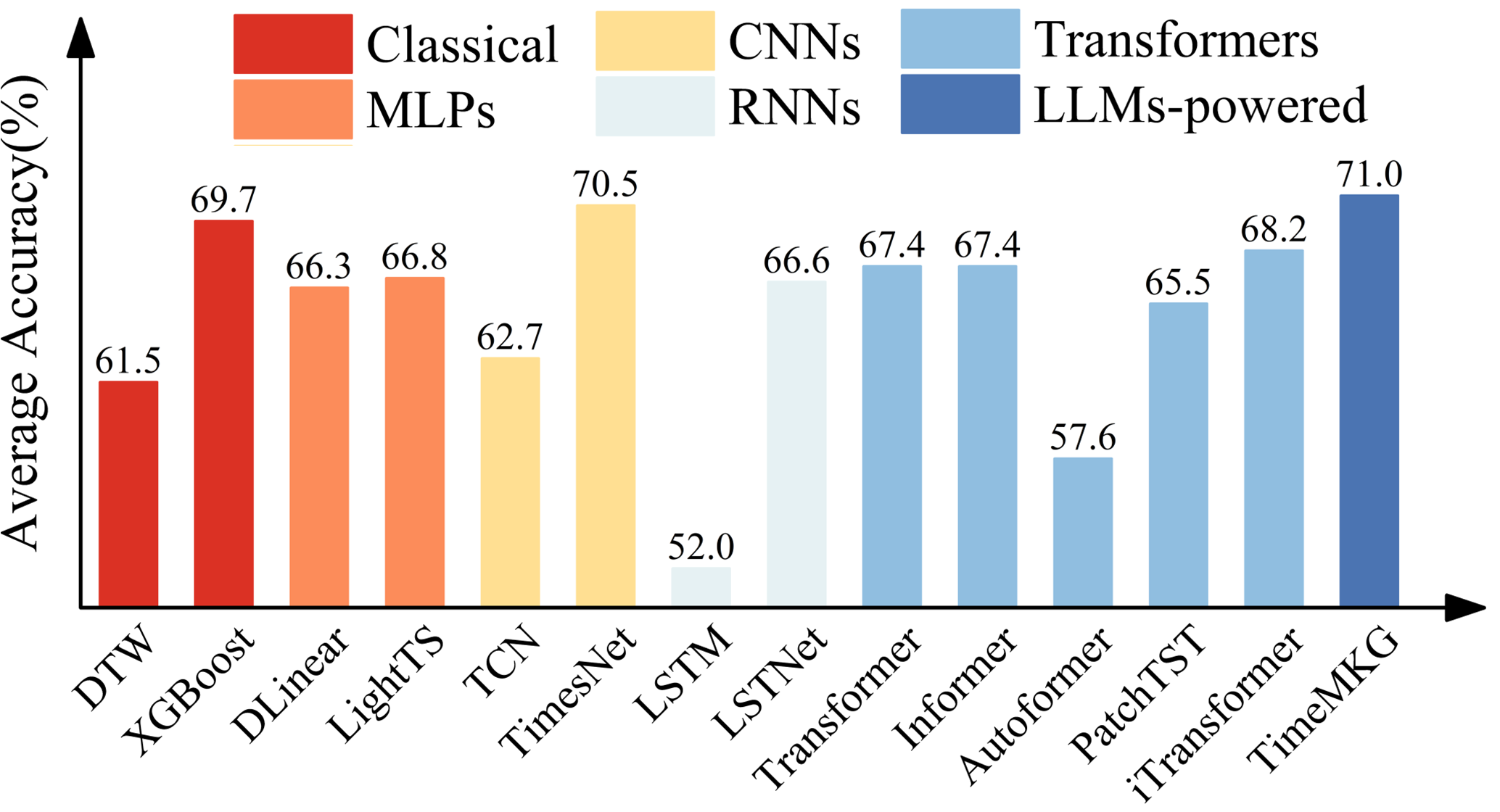}
\caption{Classification Task. The results are the average accuracy.}
\label{Figuire:classification}
\end{figure}

In Figure \ref{Figuire:classification}, in MTSD classification datasets with clearly defined header variables, TimeMKG achieves the best average predictive accuracy of 71.0\%, outperforming classical methods such as XGBoost (69.7\%) and deep learning models like TimesNet (68.2\%). Notably, TimeMKG demonstrates a substantial improvement over PatchTST (65.2\%), which performs poorly in classification tasks due to its channel-independent design that neglects inter-variable dependencies. In contrast, TimeMKG explicitly models both causal and statistical relationships among variables, equipping the model with a deeper understanding of complex multivariate entanglement in MTSD.

\begin{figure}[t]
\centering
\includegraphics[width=0.8\columnwidth]{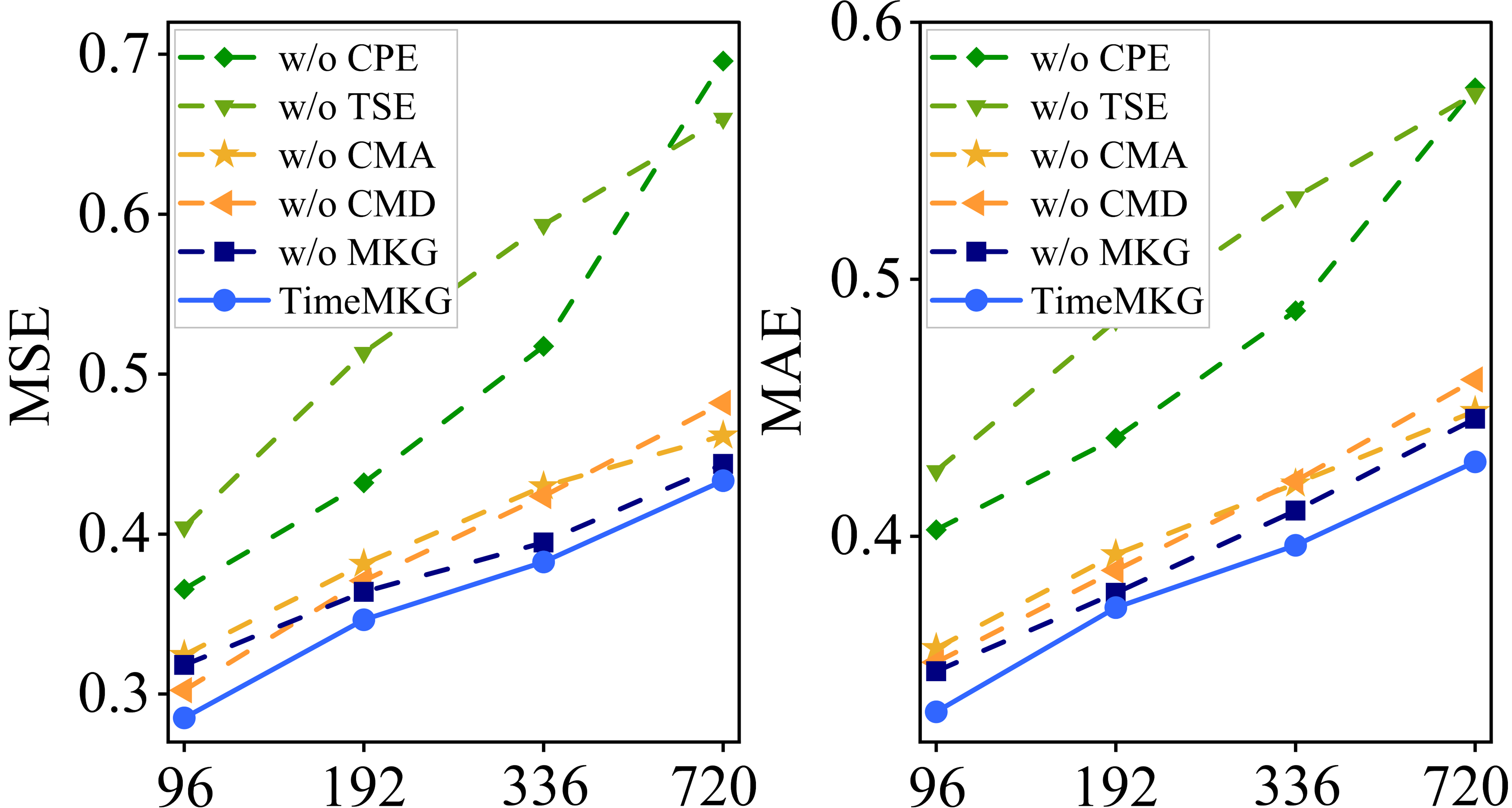}
\caption{Ablation study of model structure. The historical sequence length is fixed at \(T = 96\), and results are averaged over four subtasks in the long-term forecasting setting with prediction lengths \(L \in \{96, 192, 336, 720\}\).}
\label{Fig:ablation}
\end{figure}

\begin{figure}[t]
    \centering
    \subfloat[\(\mathit{TSEncoder}(\cdot)\).]{\includegraphics[height=3.5cm]{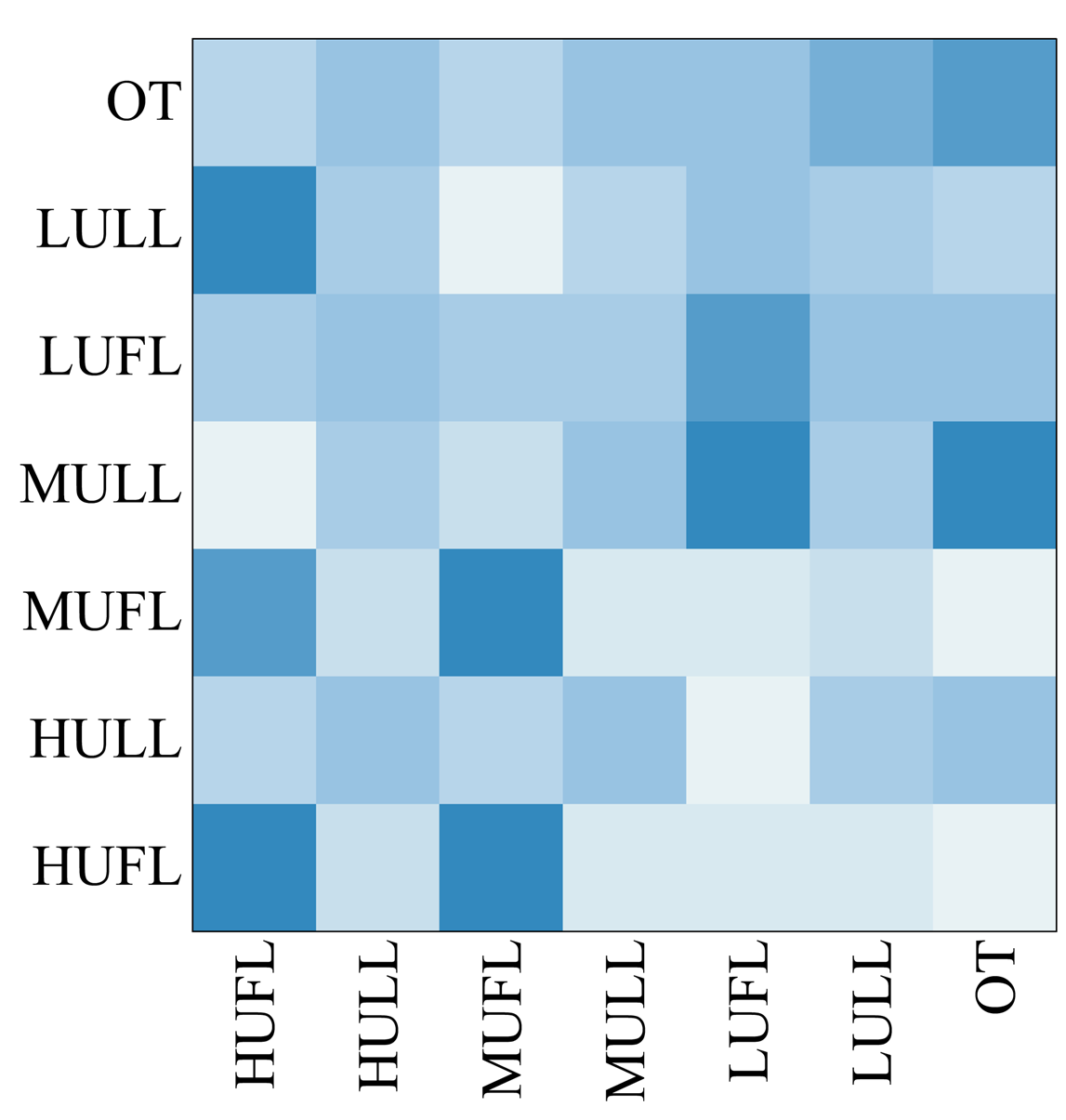}}
    \hspace{0.5cm} 
    \subfloat[\(\mathit{CPEncoder}(\cdot)\).]{\includegraphics[height=3.5cm]{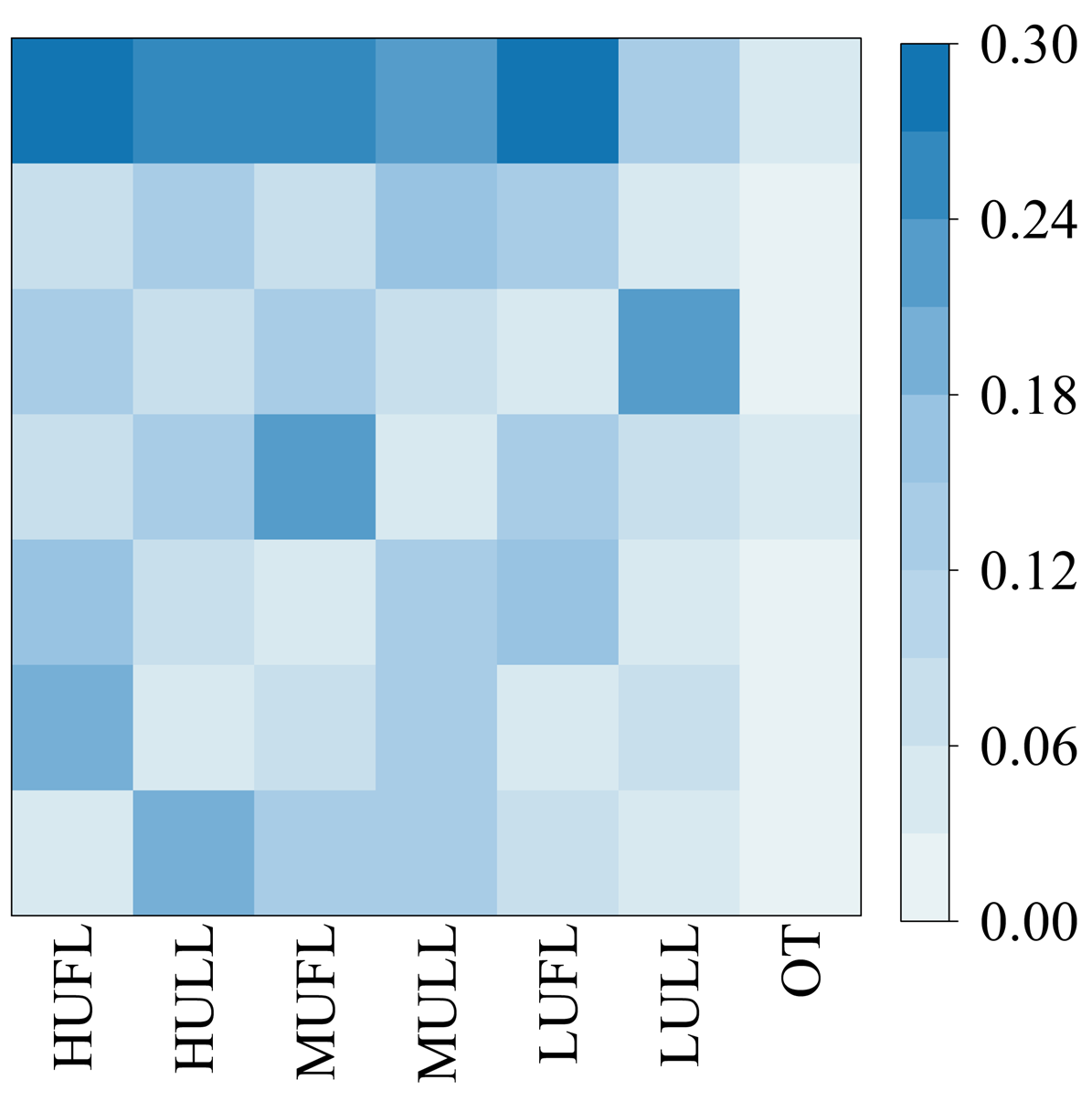}}
    \caption{Visualization of the attention score matrix \(\mathbf{S}\).}
    \label{Fig:attention score}
\end{figure}

\begin{figure}[t]
\centering
\includegraphics[width=0.8\columnwidth]{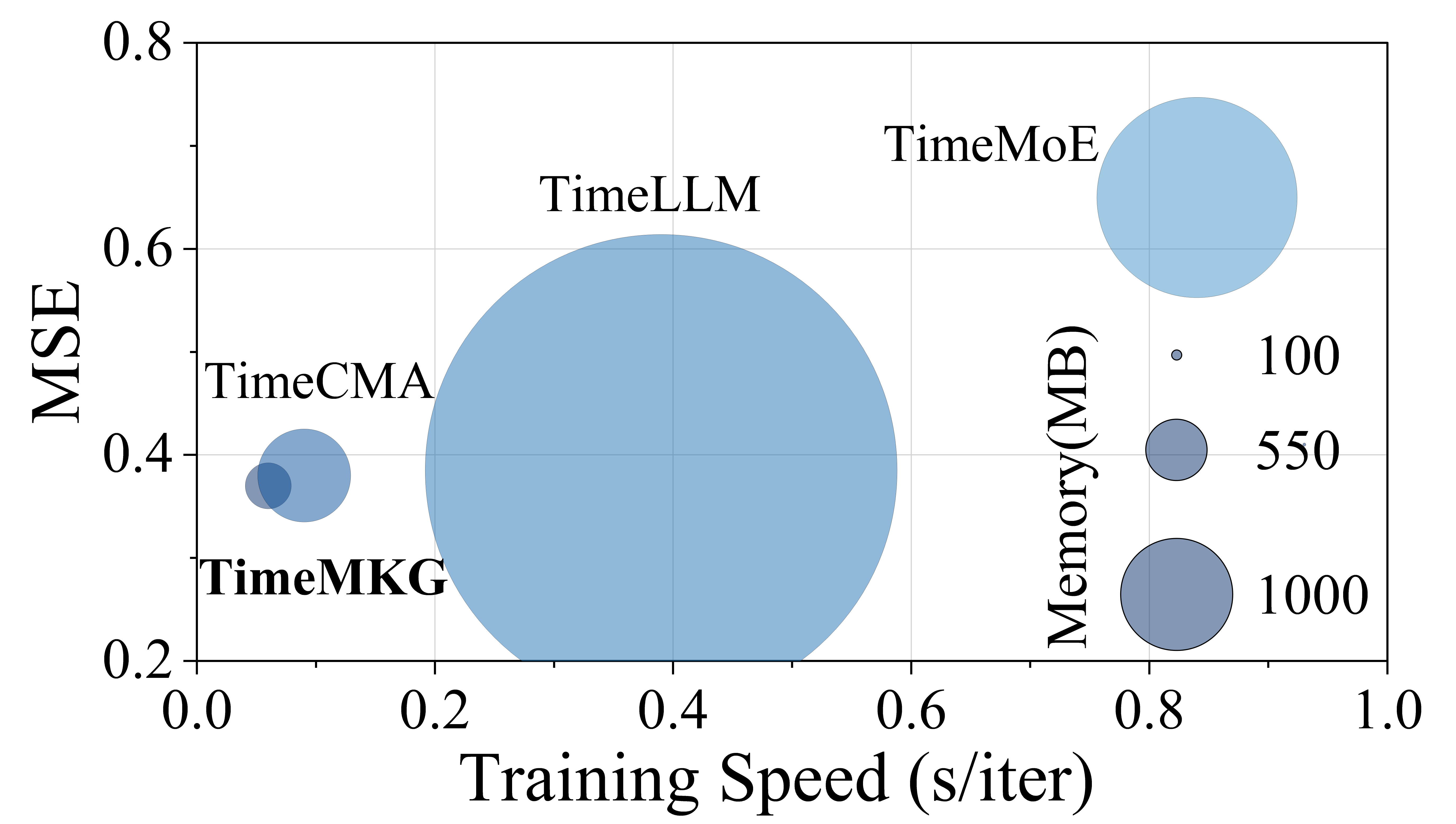}
\caption{Efficiency analysis of LLM-based methods.}
\label{Fig:param}
\end{figure}

\subsubsection{Ablation Study.}
Figure \ref{Fig:ablation} shows the ablation results on the \textit{ETT} dataset for the long-term forecasting task. The reported scores are averaged over different prediction lengths and masking ratios. The variant \textit{w\textbackslash o MKG} component queries the LLMs using knowledge as prompts, thereby bypassing the graph construction process. Results indicate that removing explicit knowledge retrieval significantly impairs the model's ability to capture causal relationships. The removal of encoder variants (\textit{w\textbackslash o CPE} and \textit{w\textbackslash o TSE}) leads to the most severe performance decline. This is because these variants directly feed the embeddings into the CMA module, resulting in the loss of learnable parameters essential for representation learning. In the variant \textit{w\textbackslash o CMA}, the outputs of the dual-branch encoders are simply concatenated. In contrast, the variant \textit{w\textbackslash o CMD} removes the decoder from the prediction module. The performance drop caused by removing the decoder is less severe than that from eliminating the cross-modal fusion structure, underscoring the crucial role of dual-modality interaction.

\subsubsection{Attention Visualization.} We visualize the attention scores of different encoders in Figures \ref{Fig:attention score}(a) and \ref{Fig:attention score}(b). Although both the  \(\mathit{TSEncoder}(\cdot)\) and the \(\mathit{CPEncoder}(\cdot)\) adopt the same Pre-LN Transformer architecture, they capture complementary inter-variable dependencies. Specifically, \(\mathit{TSEncoder}(\cdot)\) learns sample-specific dynamic correlations, reflecting statistical relationships within the data, while \(\mathit{CPEncoder}(\cdot)\) captures domain-informed static dependencies, which represent prior causal knowledge. Notably, the attention score matrix of \(\mathit{CPEncoder}(\cdot)\) closely resembles the MKG-based causal structure shown in Figure \ref{Figuire:Graphprompt}, highlighting strong associations between load variables and the target variable. This consistency suggests that \(\mathit{CPEncoder}(\cdot)\) effectively incorporates and utilizes prior knowledge embedded in the MKG.

\subsubsection{Efficiency analysis.} Figure \ref{Fig:param} presents an efficiency analysis of LLM-based methods. To ensure a fair comparison, all models are evaluated using the same batch size. Time-MoE follows its original pretraining speed and model scale settings. The results demonstrate that TimeMKG, built upon an LLMs-powered architecture, achieves significant advantages in both model size and training speed compared to LLM-backboned TimeLLM and the foundational LLM model Time-MoE. Compared to TimeCMA, TimeMKG adopts a strategy of pre-storing causal prompts, enabling faster loading of semantic representations and avoiding repeated LLM queries, which greatly reduces both runtime and memory consumption.

\section{Conclusion}
This paper proposes TimeMKG, an LLMs-powered framework for multimodal modeling of MTSD. TimeMKG leverages LLMs to extract domain-specific knowledge from textual modalities and construct a knowledge graph encoding causal relationships. A dual-modality encoder framework models both the causal representations from prompts and the statistical features from numerical time series. Cross-modality attention enables variable-level alignment and fusion between statistical time-series patterns and encoded domain knowledge for downstream tasks. Extensive experiments demonstrate the effectiveness of incorporating inter-variable causal embeddings in improving model reasoning.

\bibliography{aaai2026}

\begin{thebibliography}{40}
\providecommand{\natexlab}[1]{#1}

\bibitem[{Bai, Kolter, and Koltun(2018)}]{TCNs}
Bai, S.; Kolter, J.~Z.; and Koltun, V. 2018.
\newblock An empirical evaluation of generic convolutional and recurrent networks for sequence modeling.
\newblock \emph{arXiv preprint arXiv:1803.01271}.

\bibitem[{Berndt and Clifford(1994)}]{DTW}
Berndt, D.~J.; and Clifford, J. 1994.
\newblock Using dynamic time warping to find patterns in time series.
\newblock In \emph{Proceedings of the 3rd international conference on knowledge discovery and data mining}, 359--370.

\bibitem[{Chen and Guestrin(2016)}]{XGBoost}
Chen, T.; and Guestrin, C. 2016.
\newblock XGBoost: A Scalable Tree Boosting System.
\newblock In \emph{Proceedings of the 22nd ACM SIGKDD International Conference on Knowledge Discovery and Data Mining}, KDD '16, 785–794. New York, NY, USA: Association for Computing Machinery.
\newblock ISBN 9781450342322.

\bibitem[{Das et~al.(2023)Das, Kong, Leach, Mathur, Sen, and Yu}]{TiDE}
Das, A.; Kong, W.; Leach, A.; Mathur, S.; Sen, R.; and Yu, R. 2023.
\newblock Long-term forecasting with tide: Time-series dense encoder.
\newblock \emph{arXiv preprint arXiv:2304.08424}.

\bibitem[{Guo et~al.(2024)Guo, Xia, Yu, Ao, and Huang}]{lightrag}
Guo, Z.; Xia, L.; Yu, Y.; Ao, T.; and Huang, C. 2024.
\newblock Lightrag: Simple and fast retrieval-augmented generation.
\newblock \emph{arXiv preprint arXiv:2410.05779}.

\bibitem[{Hochreiter and Schmidhuber(1997)}]{LSTM}
Hochreiter, S.; and Schmidhuber, J. 1997.
\newblock Long short-term memory.
\newblock \emph{Neural computation}, 9(8): 1735--1780.

\bibitem[{Huang et~al.(2025)Huang, Xu, Wu, Li, and Bian}]{TimeDP}
Huang, Y.-H.; Xu, C.; Wu, Y.; Li, W.-J.; and Bian, J. 2025.
\newblock TimeDP: Learning to Generate Multi-Domain Time Series with Domain Prompts.
\newblock \emph{Proceedings of the AAAI Conference on Artificial Intelligence}, 39(17): 17520--17527.

\bibitem[{Huguet~Cabot and Navigli(2021)}]{REBEL}
Huguet~Cabot, P.-L.; and Navigli, R. 2021.
\newblock {REBEL}: Relation Extraction By End-to-end Language generation.
\newblock In Moens, M.-F.; Huang, X.; Specia, L.; and Yih, S. W.-t., eds., \emph{Findings of the Association for Computational Linguistics: EMNLP 2021}, 2370--2381. Punta Cana, Dominican Republic: Association for Computational Linguistics.

\bibitem[{Jin et~al.(2023{\natexlab{a}})Jin, Wang, Ma, Chu, Zhang, Shi, Chen, Liang, Li, Pan et~al.}]{TimeLLM}
Jin, M.; Wang, S.; Ma, L.; Chu, Z.; Zhang, J.~Y.; Shi, X.; Chen, P.-Y.; Liang, Y.; Li, Y.-F.; Pan, S.; et~al. 2023{\natexlab{a}}.
\newblock Time-llm: Time series forecasting by reprogramming large language models.
\newblock \emph{arXiv preprint arXiv:2310.01728}.

\bibitem[{Jin et~al.(2023{\natexlab{b}})Jin, Chen, Leeb, Gresele, Kamal, Lyu, Blin, Gonzalez~Adauto, Kleiman-Weiner, Sachan et~al.}]{Jin2023}
Jin, Z.; Chen, Y.; Leeb, F.; Gresele, L.; Kamal, O.; Lyu, Z.; Blin, K.; Gonzalez~Adauto, F.; Kleiman-Weiner, M.; Sachan, M.; et~al. 2023{\natexlab{b}}.
\newblock Cladder: Assessing causal reasoning in language models.
\newblock \emph{Advances in Neural Information Processing Systems}, 36: 31038--31065.

\bibitem[{Kim et~al.(2024)Kim, Kang, Kim, and Huang}]{kim2024causal}
Kim, Y.; Kang, E.; Kim, J.; and Huang, H.~H. 2024.
\newblock Causal reasoning in large language models: A knowledge graph approach.
\newblock \emph{arXiv preprint arXiv:2410.11588}.

\bibitem[{Lai et~al.(2018)Lai, Chang, Yang, and Liu}]{LSTNet}
Lai, G.; Chang, W.-C.; Yang, Y.; and Liu, H. 2018.
\newblock Modeling Long- and Short-Term Temporal Patterns with Deep Neural Networks.
\newblock In \emph{The 41st International ACM SIGIR Conference on Research \& Development in Information Retrieval}, SIGIR '18, 95–104. New York, NY, USA: Association for Computing Machinery.
\newblock ISBN 9781450356572.

\bibitem[{Li et~al.(2023)Li, Dong, Chang, Chen, Wang, Zhuang, and Yan}]{liXGboost}
Li, X.; Dong, Y.; Chang, L.; Chen, L.; Wang, G.; Zhuang, Y.; and Yan, X. 2023.
\newblock Dynamic hybrid modeling of fuel ethanol fermentation process by integrating biomass concentration XGBoost model and kinetic parameter artificial neural network model into mechanism model.
\newblock \emph{Renewable Energy}, 205: 574--582.

\bibitem[{Liu et~al.(2024{\natexlab{a}})Liu, Xu, Miao, Yang, Zhang, Long, Li, and Zhao}]{TimeCMA}
Liu, C.; Xu, Q.; Miao, H.; Yang, S.; Zhang, L.; Long, C.; Li, Z.; and Zhao, R. 2024{\natexlab{a}}.
\newblock Timecma: Towards llm-empowered time series forecasting via cross-modality alignment.
\newblock \emph{arXiv preprint arXiv:2406.01638}.

\bibitem[{Liu et~al.(2025)Liu, Meng, Gao, Mao, Cai, Yan, Chen, Bian, Shi, and Wang}]{aligning}
Liu, J.; Meng, S.; Gao, Y.; Mao, S.; Cai, P.; Yan, G.; Chen, Y.; Bian, Z.; Shi, B.; and Wang, D. 2025.
\newblock Aligning vision to language: Text-free multimodal knowledge graph construction for enhanced llms reasoning.
\newblock \emph{arXiv preprint arXiv:2503.12972}.

\bibitem[{Liu et~al.(2024{\natexlab{b}})Liu, Hu, Li, Diao, Liang, Hooi, and Zimmermann}]{UniTime}
Liu, X.; Hu, J.; Li, Y.; Diao, S.; Liang, Y.; Hooi, B.; and Zimmermann, R. 2024{\natexlab{b}}.
\newblock Unitime: A language-empowered unified model for cross-domain time series forecasting.
\newblock In \emph{Proceedings of the ACM Web Conference 2024}, 4095--4106.

\bibitem[{Liu et~al.(2023)Liu, Hu, Zhang, Wu, Wang, Ma, and Long}]{iTransformer}
Liu, Y.; Hu, T.; Zhang, H.; Wu, H.; Wang, S.; Ma, L.; and Long, M. 2023.
\newblock itransformer: Inverted transformers are effective for time series forecasting.
\newblock \emph{arXiv preprint arXiv:2310.06625}.

\bibitem[{Mameche et~al.(2025)Mameche, Cornanguer, Ninad, and Vreeken}]{casualtime}
Mameche, S.; Cornanguer, L.; Ninad, U.; and Vreeken, J. 2025.
\newblock SPACETIME: Causal Discovery from Non-Stationary Time Series.
\newblock \emph{Proceedings of the AAAI Conference on Artificial Intelligence}, 39(18): 19405--19413.

\bibitem[{Nie et~al.(2022)Nie, Nguyen, Sinthong, and Kalagnanam}]{PatchTST}
Nie, Y.; Nguyen, N.~H.; Sinthong, P.; and Kalagnanam, J. 2022.
\newblock A time series is worth 64 words: Long-term forecasting with transformers.
\newblock \emph{arXiv preprint arXiv:2211.14730}.

\bibitem[{Oreshkin et~al.(2019)Oreshkin, Carpov, Chapados, and Bengio}]{N-BEATS}
Oreshkin, B.~N.; Carpov, D.; Chapados, N.; and Bengio, Y. 2019.
\newblock N-BEATS: Neural basis expansion analysis for interpretable time series forecasting.
\newblock \emph{arXiv preprint arXiv:1905.10437}.

\bibitem[{Qin et~al.(2017)Qin, Song, Chen, Cheng, Jiang, and Cottrell}]{DARNN}
Qin, Y.; Song, D.; Chen, H.; Cheng, W.; Jiang, G.; and Cottrell, G. 2017.
\newblock A dual-stage attention-based recurrent neural network for time series prediction.
\newblock \emph{arXiv preprint arXiv:1704.02971}.

\bibitem[{Salinas et~al.(2020)Salinas, Flunkert, Gasthaus, and Januschowski}]{deepar}
Salinas, D.; Flunkert, V.; Gasthaus, J.; and Januschowski, T. 2020.
\newblock DeepAR: Probabilistic forecasting with autoregressive recurrent networks.
\newblock \emph{International journal of forecasting}, 36(3): 1181--1191.

\bibitem[{Seong, Lee, and Chae(2024)}]{seong2024self}
Seong, E.; Lee, H.; and Chae, D.-K. 2024.
\newblock Self-supervised framework based on subject-wise clustering for human subject time series data.
\newblock In \emph{Proceedings of the AAAI Conference on Artificial Intelligence}, volume~38, 22341--22349.

\bibitem[{Sezer, Gudelek, and Ozbayoglu(2020)}]{finance}
Sezer, O.~B.; Gudelek, M.~U.; and Ozbayoglu, A.~M. 2020.
\newblock Financial time series forecasting with deep learning : A systematic literature review: 2005–2019.
\newblock \emph{Applied Soft Computing}, 90: 106181.

\bibitem[{Shi et~al.(2024)Shi, Wang, Nie, Li, Ye, Wen, and Jin}]{Time-MoE}
Shi, X.; Wang, S.; Nie, Y.; Li, D.; Ye, Z.; Wen, Q.; and Jin, M. 2024.
\newblock Time-moe: Billion-scale time series foundation models with mixture of experts.
\newblock \emph{arXiv preprint arXiv:2409.16040}.

\bibitem[{Susanti and F{\"a}rber(2024)}]{susanti2024knowledge}
Susanti, Y.; and F{\"a}rber, M. 2024.
\newblock Knowledge graph structure as prompt: improving small language models capabilities for knowledge-based causal discovery.
\newblock In \emph{International Semantic Web Conference}, 87--106. Springer.

\bibitem[{Tan et~al.(2024)Tan, Merrill, Gupta, Althoff, and Hartvigsen}]{areuse}
Tan, M.; Merrill, M.; Gupta, V.; Althoff, T.; and Hartvigsen, T. 2024.
\newblock Are language models actually useful for time series forecasting?
\newblock \emph{Advances in Neural Information Processing Systems}, 37: 60162--60191.

\bibitem[{Wang et~al.(2025)Wang, Qi, Wang, Sun, Zhuang, Wu, Zhang, and Liao}]{ChatTime}
Wang, C.; Qi, Q.; Wang, J.; Sun, H.; Zhuang, Z.; Wu, J.; Zhang, L.; and Liao, J. 2025.
\newblock ChatTime: A Unified Multimodal Time Series Foundation Model Bridging Numerical and Textual Data.
\newblock \emph{Proceedings of the AAAI Conference on Artificial Intelligence}, 39(12): 12694--12702.

\bibitem[{Wang, Zhu, and He(2024)}]{Tii}
Wang, P.-F.; Zhu, Q.-X.; and He, Y.-L. 2024.
\newblock Novel Multiscale Trend Decomposition LSTM Based on Feature Selection for Industrial Soft Sensing.
\newblock \emph{IEEE Transactions on Industrial Informatics}, 20(12): 14249--14256.

\bibitem[{Wang et~al.(2024{\natexlab{a}})Wang, Wu, Dong, Liu, Qiu, Zhang, Wang, and Long}]{TimeXer}
Wang, Y.; Wu, H.; Dong, J.; Liu, Y.; Qiu, Y.; Zhang, H.; Wang, J.; and Long, M. 2024{\natexlab{a}}.
\newblock TimeXer: Empowering Transformers for Time Series Forecasting with Exogenous Variables.
\newblock \emph{ArXiv}, abs/2402.19072.

\bibitem[{Wang et~al.(2024{\natexlab{b}})Wang, Xu, Yang, Wu, Li, Xie, and Chen}]{FCSTGNN}
Wang, Y.; Xu, Y.; Yang, J.; Wu, M.; Li, X.; Xie, L.; and Chen, Z. 2024{\natexlab{b}}.
\newblock Fully-connected spatial-temporal graph for multivariate time-series data.
\newblock In \emph{Proceedings of the AAAI conference on artificial intelligence}, volume~38, 15715--15724.

\bibitem[{Woo et~al.(2024)Woo, Liu, Kumar, Xiong, Savarese, and Sahoo}]{Moirai}
Woo, G.; Liu, C.; Kumar, A.; Xiong, C.; Savarese, S.; and Sahoo, D. 2024.
\newblock Unified Training of Universal Time Series Forecasting Transformers.
\newblock \emph{ArXiv}, abs/2402.02592.

\bibitem[{Wu et~al.(2022)Wu, Hu, Liu, Zhou, Wang, and Long}]{Timesnet}
Wu, H.; Hu, T.; Liu, Y.; Zhou, H.; Wang, J.; and Long, M. 2022.
\newblock Timesnet: Temporal 2d-variation modeling for general time series analysis.
\newblock \emph{arXiv preprint arXiv:2210.02186}.

\bibitem[{Yang et~al.(2025)Yang, Li, Yang, Zhang, Hui, Zheng, Yu, Gao, Huang, Lv et~al.}]{qwen3}
Yang, A.; Li, A.; Yang, B.; Zhang, B.; Hui, B.; Zheng, B.; Yu, B.; Gao, C.; Huang, C.; Lv, C.; et~al. 2025.
\newblock Qwen3 technical report.
\newblock \emph{arXiv preprint arXiv:2505.09388}.

\bibitem[{Zeng et~al.(2023)Zeng, Chen, Zhang, and Xu}]{Dlinear}
Zeng, A.; Chen, M.; Zhang, L.; and Xu, Q. 2023.
\newblock Are transformers effective for time series forecasting?
\newblock In \emph{Proceedings of the AAAI conference on artificial intelligence}, volume~37, 11121--11128.

\bibitem[{Zhang et~al.(2022)Zhang, Zhang, Cao, Bian, Yi, Zheng, and Li}]{lightTS}
Zhang, T.; Zhang, Y.; Cao, W.; Bian, J.; Yi, X.; Zheng, S.; and Li, J. 2022.
\newblock Less Is More: Fast Multivariate Time Series Forecasting with Light Sampling-oriented MLP Structures.
\newblock \emph{ArXiv}, abs/2207.01186.

\bibitem[{Zhao et~al.(2025)Zhao, Wang, Wen, Wang, Yu, and Wang}]{STEM-LTS}
Zhao, Z.; Wang, P.; Wen, H.; Wang, S.; Yu, L.; and Wang, Y. 2025.
\newblock STEM-LTS: Integrating Semantic-Temporal Dynamics in LLM-driven Time Series Analysis.
\newblock \emph{Proceedings of the AAAI Conference on Artificial Intelligence}, 39(21): 22858--22866.

\bibitem[{Zheng et~al.(2020)Zheng, Fan, Wang, and Qi}]{GMAN}
Zheng, C.; Fan, X.; Wang, C.; and Qi, J. 2020.
\newblock Gman: A graph multi-attention network for traffic prediction.
\newblock In \emph{Proceedings of the AAAI conference on artificial intelligence}, volume~34, 1234--1241.

\bibitem[{Zhou et~al.(2021)Zhou, Zhang, Peng, Zhang, Li, Xiong, and Zhang}]{informer}
Zhou, H.; Zhang, S.; Peng, J.; Zhang, S.; Li, J.; Xiong, H.; and Zhang, W. 2021.
\newblock Informer: Beyond efficient transformer for long sequence time-series forecasting.
\newblock In \emph{Proceedings of the AAAI conference on artificial intelligence}, volume~35, 11106--11115.

\bibitem[{Zhou et~al.(2022)Zhou, Ma, Wen, Wang, Sun, and Jin}]{fedformer}
Zhou, T.; Ma, Z.; Wen, Q.; Wang, X.; Sun, L.; and Jin, R. 2022.
\newblock Fedformer: Frequency enhanced decomposed transformer for long-term series forecasting.
\newblock In \emph{International conference on machine learning}, 27268--27286. PMLR.

\end{thebibliography}

\end{document}